\documentclass[12pt]{article}
\usepackage[utf8]{inputenc}
\usepackage[letterpaper, margin=2.5cm, top=2.5cm, bottom=2.5cm]{geometry}
\usepackage{amssymb}
\usepackage{amsmath}
\usepackage{graphicx}
\usepackage{mathtools}
\usepackage{xcolor}
\usepackage{pdfpages}
\usepackage{multirow}
\usepackage{dsfont}
\usepackage{bbm}
\usepackage{amsfonts}
\usepackage{multicol}
\usepackage{bm}
\usepackage{dirtytalk}
\usepackage{abstract}
\usepackage[round]{natbib}
\usepackage{hyperref}
\definecolor{citeblue}{rgb}{0.05, 0.42, 0.87}
\hypersetup{colorlinks, citecolor=citeblue}
\usepackage[inline]{enumitem}
\usepackage{footmisc}
\usepackage{mwe}
\usepackage{subcaption}

\DeclareMathOperator*{\argmax}{arg\,max}
\DeclareMathOperator*{\argmin}{arg\,min}

\graphicspath{{figs/}}
\usepackage[english]{babel}
\usepackage{enumitem}
\usepackage{listings}
\usepackage{filecontents}
\usepackage{verbatim}
\usepackage{eurosym}
\usepackage[export]{adjustbox}

\begin{document}

\pagenumbering{gobble}


\begin{titlepage}

\newcommand{\HRule}{\rule{\linewidth}{0.5mm}} 

\center 
 

\textsc{\LARGE University of Cambridge}\\[1.5cm] 
\textsc{\Large Department of Engineering}\\[0.5cm] 


\HRule \\[0.4cm]
{ \huge \bfseries Amortised Inference in Bayesian Neural Networks}\\[0.4cm] 
\HRule \\[1.5cm]
 

\begin{minipage}{0.4\textwidth}
\begin{flushleft} \large
\emph{Author:}\\
Tommy Rochussen \\
Downing College \\
\end{flushleft}
\end{minipage}
~
\begin{minipage}{0.4\textwidth}
\begin{flushright} \large
\emph{Supervisors:} \\
Dr. Adrian Weller \\
Matthew Ashman \\
\end{flushright}
\end{minipage}\\[2cm]



{\large June 2023}\\[2cm] 


\includegraphics[width=0.2\textwidth]{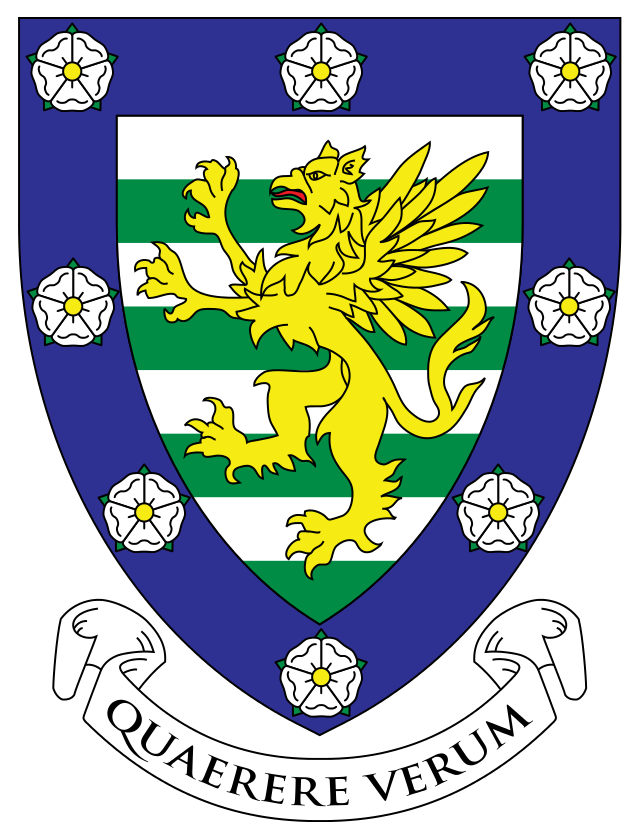}\\[1cm] 
 

\vfill 

\end{titlepage}

\begin{abstract}
\normalsize \textit{Meta-learning} is a framework in which machine learning models train over a set of datasets in order to produce predictions on new datasets at test time. Probabilistic meta-learning has received an abundance of attention from the research community in recent years, but a problem shared by many existing probabilistic meta-models is that they require a very large number of datasets in order to produce high-quality predictions with well-calibrated uncertainty estimates. In many applications, however, such quantities of data are simply not available. 

In this dissertation we present a significantly more data-efficient approach to probabilistic meta-learning through per-datapoint amortisation of inference in Bayesian neural networks, introducing the Amortised Pseudo-Observation Variational Inference Bayesian Neural Network (APOVI-BNN). First, we show that the approximate posteriors obtained under our amortised scheme are of similar or better quality to those obtained through traditional variational inference, despite the fact that the amortised inference is performed in a single forward pass. We then discuss how the APOVI-BNN may be viewed as a new member of the neural process family, motivating the use of neural process training objectives for potentially better predictive performance on complex problems as a result. Finally, we assess the predictive performance of the APOVI-BNN against other probabilistic meta-models in both a one-dimensional regression problem and in a significantly more complex image completion setting. In both cases, when the amount of training data is limited, our model is the best in its class.
\end{abstract}

\newpage

\begin{centering}
\section*{Declaration}
\end{centering}
\textit{I, Thomas Nicholas Rochussen, hereby declare that, except where specifically indicated, the work submitted herein is my own original work. All sources used have been duly acknowledged.}

\newpage

\tableofcontents

\clearpage
\pagenumbering{arabic}

\section{Introduction}

Machine learning models that produce predictions with accurate uncertainty estimates are of vital importance across a range of domains. In many applications, such predictive distributions are required for multiple datasets which are related in some way, often by being of the same form. A common example is that of a machine learning system for personalised health, in which each patient corresponds to a distinct dataset. Naturally we would like to avoid having to train a model from scratch for each dataset, but rather use one model that can make predictions for each new dataset at test time. To do this, the model would need to \say{learn how to learn}; during training it would learn to make predictions on unseen datasets, as opposed to unseen datapoints as in a regular model. Such a process is known as meta-learning. For parametric models, meta-learning is the process of finding a set of model parameters that are shared across all datasets (referred to as \textit{tasks} in the meta-learning literature), which allow the model to learn, at test time, any task-specific parameters needed for prediction.

Meta-learning has become a popular topic in the field over the past few years, with frameworks such as neural processes (NPs) \citep{garnelo2018conditional, garnelo2018neural} becoming particularly prevalent. While many methods work well on very large sets of datasets (meta-datasets), generally their performance is poor when the number of datasets is limited. Arguably this is because of two reasons:
\begin{enumerate*}
    \item \textit{they rely on shared model parameters overfitting to the meta-dataset}
    \item \textit{existing posterior distribution approximations do not exploit model structure for efficient learning.}
\end{enumerate*}
An obvious remedy for the first issue is to perform Bayesian inference over the shared parameters---not just the task-specific ones. The second problem is more difficult to solve since it would involve devising a bespoke approximate posterior distribution for a given model architecture that builds in as much structural information as possible.

Recently, \cite{ober2021global} posited a new variational approximate posterior for Bayesian neural networks (BNNs) that uses a set of trainable \textit{global inducing points} and whose construction mimics that of the true posterior distribution. Crucially, their approximate posterior allows for network weight correlations across network layers to be modelled. The key insight of this project is that instead of learning the set of global inducing locations as \citeauthor{ober2021global} do, we can set the available datapoints as the inducing locations, allowing the approximate posterior distribution to be decomposed into a product of per-datapoint approximate likelihoods and priors. Such a decomposition enables us to obtain the parameters of each approximate likelihood by passing the corresponding datapoint through a secondary inference network (typically an MLP)---a process known as amortised inference. The parameters of the secondary networks are then the variational parameters, and are found by optimising the evidence lower bound (ELBO) across a meta-dataset. After training, the secondary networks will have learned to perform approximate Bayesian inference in the primary network on unseen datasets, and as such the framework as a whole addresses both of the limitations detailed above. The core contributions of this project are outlined as follows:
\begin{enumerate}
    \item \textbf{A way to perform amortised variational inference in BNNs}
    \item \textbf{The connection between the proposed model and the NP family}
    \item \textbf{Evidence that the proposed model outperforms existing models on small-scale meta-learning problems}
\end{enumerate}

It turns out that there are distinct similarities between our method and members of the latent NP family \citep{garnelo2018neural}, inspiring the application of NP training methods to our model. It is then natural to assess the performance of our model across problems that are typical in the NP literature; one-dimensional toy regressions, and image completion \citep{gordon2020convolutional}. In both of these settings, when the number of datasets across which the metamodels train is limited, we find that our method performs better than existing probabilistic meta-models.

\section{Background}

\subsection{Approximate Bayesian Inference}
Given a statistical model $p(\mathcal{D}|\boldsymbol{\theta})$ for observed data $\mathcal{D}$ parameterised by a set of potentially multidimensional model parameters $\boldsymbol{\theta}$, and a prior distribution over the model parameters $p(\boldsymbol{\theta})$, the posterior distribution over model parameters $p(\boldsymbol{\theta}|\mathcal{D})$ is found by applying Bayes' theorem:
\begin{equation*}
    p(\boldsymbol{\theta}|\mathcal{D}) = \frac{p(\mathcal{D}|\boldsymbol{\theta})p(\boldsymbol{\theta})}{p(\mathcal{D})}.
\end{equation*}The process of computing posterior distributions in this way is known as performing Bayesian inference \citep{bishop2007}. The constant-term denominator, the marginal likelihood, is found by computing
\begin{equation*}
    p(\mathcal{D}) = \int p(\mathcal{D}|\boldsymbol{\theta})p(\boldsymbol{\theta}) \mathrm{d}\boldsymbol{\theta}.
\end{equation*}Unfortunately, for almost all applications of practical interest there is not an analytic solution to this integral, and numerical integration methods typically suffer at the hands of the \textit{curse of dimensionality} \citep{bellman1966dynamic} for models with many parameters. As a result, we are left unable to compute the scaling factor that converts the joint distribution $p(\mathcal{D}, \boldsymbol{\theta})$ to the desired posterior distribution. Instead, we must resort to approximating the posterior distribution in some way.

There are two major classes of approximate inference scheme; deterministic approximations and stochastic approximations. Deterministic approximations entail selecting a tractable distribution that mimics the true posterior distribution in some way, before proceeding with that distribution in place of the exact posterior. One deterministic approximation of significance is the Laplace approximation (\citealt{doi:10.1080/01621459.1986.10478240}, \citealt{MacKay2003}), in which a Gaussian distribution is fitted to the mode of the true posterior using the mode location and Hessian thereof to find the Gaussian mean and covariance parameters respectively, providing a good approximation to the posterior near its mode. Another prominent deterministic approximation method is expectation propagation (EP) \citep{minka2013expectation}, an algorithm that iteratively updates the factors of the factorisable approximation by minimising the (backward) Kullback-Leibler (KL) divergence between the approximation factor and the true posterior conditioned on the remaining factors, subject to known moment constraints.

Stochastic approximations involve producing samples from known distributions to estimate unknown quantities, and the most notable subclass of stochastic approximation is the Markov Chain Monte Carlo (MCMC) \citep{Robert_2011} family of methods. In an MCMC scheme, we produce samples from a Markov Chain that has been carefully crafted to have its stationary distribution equal to the true posterior of interest, and as such we can obtain samples (albeit dependent ones) from the posterior distribution. While this appears not to solve the original problem of approximating the exact posterior distribution, it is important to note that we are in general only interested in the posterior distribution in order to compute the Bayesian predictive distribution over an unseen test datapoint $\mathcal{D}^*$
\begin{equation*}
    p(\mathcal{D}^*|\mathcal{D}) = \int p(\mathcal{D}^*|\boldsymbol{\theta}) p(\boldsymbol{\theta}|\mathcal{D})\mathrm{d}\boldsymbol{\theta}
\end{equation*}which, if we have samples from the posterior distribution, can be easily approximated through simple Monte Carlo integration:
\begin{equation*}
        p(\mathcal{D}^*|\mathcal{D}) \simeq \frac{1}{M}\sum_{m=1}^M p(\mathcal{D}^*|\boldsymbol{\theta}^{(m)}), \quad \boldsymbol{\theta}^{(m)}\sim p(\boldsymbol{\theta}|\mathcal{D}).
\end{equation*}

MCMC methods typically provide samples that are consistent with the true posterior distribution and so predictive distributions aquired through MCMC methods can often be treated as exact Bayesian predictive distributions, however they are generally slow to converge and perform poorly in high dimensions. By contrast, although they might not recover the exact posterior distribution, deterministic approximations tend to be \textit{significantly} faster to compute and good deterministic approximations can be very accurate indeed, and as such we are motivated to pursue good deterministic posterior approximations in BNNs. While there are many subclasses of deterministic approximation, the most fruitful for the application of BNNs in recent times is surely that of variational inference.

\subsection{Variational Inference}
The central idea behind variational inference (VI) (\citealt{journals/ml/JordanGJS99}, \citealt{Blei_2017}) is to restrict the approximate posterior distribution $q(\boldsymbol{\theta})$ to be a member of a family of tractable distributions, and then find the member of the family that best approximates the true posterior distribution via optimisation. In particular, the optimisation seeks to minimise the (forward) KL divergence between the true and approximate posteriors such that
\begin{equation*}
    q^*(\boldsymbol{\theta}) = \argmin_{q}\text{KL}\left[q(\boldsymbol{\theta})\|p(\boldsymbol{\theta}|\mathcal{D})\right].
\end{equation*}Since the true posterior distribution is unknown, we cannot minimise the KL divergence directly but rather we maximise an equivalent objective known as the evidence lower bound (ELBO) given by
\begin{align}
    \mathcal{L}_\text{ELBO} &= \log p(\mathcal{D}) - \text{KL}\left[q(\boldsymbol{\theta})\|p(\boldsymbol{\theta}|\mathcal{D})\right] \\
    &= \int q(\boldsymbol{\theta})\log\frac{p(\boldsymbol{\theta},\mathcal{D})}{q(\boldsymbol{\theta})}\mathrm{d}\boldsymbol{\theta} \\
    &= \mathbb{E}_{q(\boldsymbol{\theta})}\left(\log p(\mathcal{D}||\boldsymbol{\theta})\right) - \text{KL}\left[q(\boldsymbol{\theta})\|p(\boldsymbol{\theta})\right].
\end{align}Note that the expression in (2) is the negative of a term known as the \textit{variational free energy}, which is used in the expectation maximisation algorithm (\citealt{10.2307/2984875}, \citealt{mclachlant97}). For a given model specification, the marginal likelihood is simply a constant term and so maximising the ELBO is exactly equivalent to minimising the forward KL divergence as seen in (1). It is an added benefit of VI that the ELBO also provides us with a lower bound for the (log) marginal likelihood, which can be utilised in the evidence framework for model selection \citep{lotfi2023bayesian}. This lower bound guarantee follows from the non-negativity property of the KL divergence. We are provided with a good intuition behind maximising the ELBO from (3); we maximise the fit to the data (first term) while remaining somewhat faithful to the prior (second term)---the usual Bayesian Occam's Razor tradeoff. Generally the approximate posterior distribution is parameterised by a set of parameters $\phi$ referred to as \textit{variational parameters}, and so optimisation is usually carried out with respect to these:
\begin{equation*}
    \phi^* = \argmax_\phi \mathcal{L}_\text{ELBO}
\end{equation*}

\subsection{Variational Inference in BNNs}

Throughout this section and beyond, let $\mathcal{D} = (\mathbf{X}, \mathbf{y})$ denote a dataset consisting of inputs $\mathbf{X}$ and outputs $\mathbf{y}$, where $\mathbf{X} \in \mathbb{R}^{N \times D}$ denotes $N$ $D$-dimensional inputs $\{\mathbf{x}_n\}_{n=1}^N$ and $\mathbf{y} \in \mathbb{R}^{N \times P}$ denotes $N$ $P$-dimensional outputs $\{\mathbf{y}_n\}_{n=1}^N$. Let $\mathbf{W} = \{\mathbf{W}^{\ell}\}_{l=1}^L$ denote the weights of a neural network with $L$ layers, such that\ $\mathbf{W}^{\ell} \in \mathbb{R}^{D^{\ell - 1} \times D^{\ell}}$ where $D^{\ell}$ is the number of units in the $\ell$-th hidden layer, and let $\psi(\cdot)$ denote the element-wise activation function acting between layers. For input datapoint $n$, let $\mathbf{F}^\ell_{n} = \psi(\mathbf{F}^{\ell-1}_{n})\mathbf{W}^\ell$ denote the output of layer $\ell\in\{2,...,L\}$ of the network and $\mathbf{F}^1_{n} = \mathbf{x}_n\mathbf{W}^1$ denote the output of the first layer. Note that we are not explicit about using network biases since they may be concatenated into the weights. Finally, let $\phi$ denote the set of variational parameters. Note that where $\boldsymbol{\theta}$ was used in previous sections to denote model parameters, in this section and beyond we use $\mathbf{W}$ instead since we are dealing with model parameters that are network weights.

For all BNNs in this section and beyond we assume independent, zero-mean Gaussian priors over model weights with variance $\sigma_p^2$
\begin{equation*}
    p(w_{ij}^\ell) = \prod_{\ell=1}^L\prod_{i=1}^{D^{\ell-1}}\prod_{j=1}^{D^\ell}\mathcal{N}\left(w_{ij}^\ell;0,\sigma_p^2\right), \quad D^0 = D
\end{equation*}and a Gaussian likelihood with observation covariance matrix $\boldsymbol{\Sigma_{noise}}\in\mathbb{R}^{P\times P}$ 
\begin{equation*}
    p(\mathcal{D}|\mathbf{W}) = p(\mathbf{y}|\mathbf{X}, \mathbf{W}) = \prod_n\mathcal{N}\left(\mathbf{y}_n;\mathbf{F}^L_{n}, \boldsymbol{\Sigma}_{noise}\right)
\end{equation*}Regardless of the form of the approximate posterior in BNNs, we cannot compute the expected log-likelihood of the data---the first term in the ELBO---in closed form. However, we can estimate the expectation by Monte Carlo integration
\begin{equation*}
    \mathbb{E}_{q_\phi(\mathbf{W})}\left(\log p(\mathbf{y}|\mathbf{X},\mathbf{W}) \right) \simeq \frac{1}{M}\sum_{m=1}^M\log p(\mathbf{y}|\mathbf{X},\mathbf{W}^{(m)}), \quad \mathbf{W}^{(m)} \sim q_\phi(\mathbf{W}).
\end{equation*}Unfortunately, this stochastic estimate prevents us from propagating gradients through the network---something that is essential for optimisation by backpropagation \citep{rumelhart1986learning}, but we can use the \textit{reparameterisation trick} (\citealt{Kingma2014}, \citealt{blundell2015weight}) to sidestep this issue. For each network weight $w_{ij}^{\ell}$, we reparameterise it as
\begin{equation*}
    w_{ij}^{\ell} = \mu_{ij}^{\ell} + \sigma_{ij}^{\ell} \epsilon_{ij}^{\ell}, \quad\text{where } \epsilon_{ij}^{\ell} \sim \mathcal{N}(\epsilon_{ij}^{\ell}; 0, 1)
\end{equation*}so that each weight can be treated as a linear combination of a deterministic weight mean and a stochastic noise term, which is differentiable. With this technology, we are now armed to train a BNN through VI; all that remains to do is to select a family of approximate posterior distributions over which to optimise.

Before exploring different families of variational posterior, it is worth mentioning some of the more popular non-VI approaches to BNNs for completeness. \cite{gal2016dropout} reinterpret dropout \citep{srivastava2014dropout} from a Bayesian perspective, obtaining model uncertainty estimates by randomly suppressing neuron outputs. It is both a simple and computationally efficient approach and one that scales well to very large and complex architectures as a result, however the quality of the uncertainty estimates are sensitive to choices of architecture and training procedure \citep{verdoja2021notes}. The de facto MCMC-based approach to BNNs is Hamiltonian Monte Carlo (HMC) \citep{NIPS1992_f29c21d4} combined with the No-U-Turn Sampler (NUTS) \citep{hoffman2011nouturn}. HMC/NUTS explores the posterior distribution more efficiently than existing MCMC approaches, however it does not scale well to high dimensions, and so cannot be used in large BNNs with many parameters.

Returning to the VI framework, there are a number of notable variational approximate posteriors for use in BNNs. Normalising flows (\citealt{rezende2016variational}, \citealt{louizos2017multiplicative}) are a series of invertible transformations which can be used to map from a simple distribution such as a Gaussian to the posterior distribution. While the ability of normalising flows to capture intricate dependencies and nonlinearities allows for flexible posterior approximations, their performance depends heavily on the choice of transformation. Moreover, normalising flows require the transforms to be invertible---something that can be somewhat restrictive. However, by far the most common type of approximate posterior is the family of Gaussian distributions.

\subsubsection{Mean Field Variational Inference in BNNs}
\begin{figure}[h]
    \centering
    \includegraphics[width=0.7\textwidth]{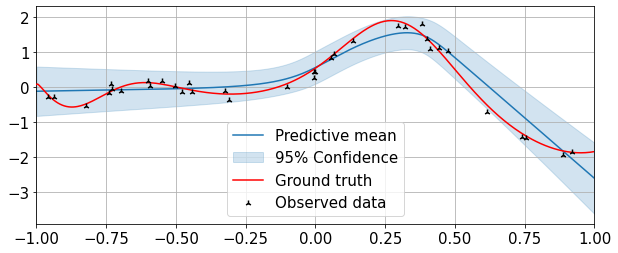}
    \caption{MFVI BNN predictive distribution on a toy regression example. The ground truth function was assembled using a highly nonlinear combination of sinusoidal, exponential, and polynomial terms, followed by whitening the function on the $[-1, 1]$ interval. The samples were generated by sampling uniformly on the $[-1, 1]$ interval and corrupting the obtained function outputs with Gaussian noise. The predictive distribution was approximated by moment-matching a Gaussian to the sample mean and variance over 100 BNN prediction samples.}
    \label{fig:mfvi-bnn}
\end{figure}

One of the simplest and most popular variational approximations for BNNs is the mean-field variational inference (MFVI) Gaussian approximation (\citealt{10.1145/168304.168306}, \citealt{blundell2015weight}), in which each network weight is assumed to follow a univariate Gaussian distribution such that the full approximate posterior is given by
\begin{equation*}
    q(\mathbf{W}) = \prod_{\ell=1}^L\prod_{i=1}^{D^{\ell-1}}\prod_{j=1}^{D^\ell}\mathcal{N}\left(w_{ij}^\ell;\mu_{ij}^\ell,{\sigma_{ij}^\ell}^2\right)
\end{equation*}The means and variances for the network weights are variational parameters, and so they are found by optimising the ELBO as above. Although this method has the advantages that it is simple to understand and straightforward to implement, it cannot model dependencies between any weights. That shortcoming expresses itself in the form of underfitting the data, a pathology that can be seen in the example in Figure \ref{fig:mfvi-bnn}. 

There have been proposed Gaussian approximate posteriors that use full covariance matrices to characterise the dependencies between weights in a given layer (\citealt{pmlr-v48-louizos16}, \citealt{ritter2018a}), however almost none are capable of modelling weight correlations \textit{across network layers}.

\subsubsection{Pseudo-Observation Variational Inference in BNNs}

\cite{ober2021global} present what they refer to as the Global Inducing Point Variational posterior approximation for BNNs, in which one of their primary goals was to allow for weight correlations to be modelled both between weights in a given layer, as well as across layers. To do this, they observed that the true posterior distribution for a BNN can be decomposed into a product of layerwise conditional posterior distributions using the chain rule of probability
\begin{align*}
    p(\mathbf{W}|\mathcal{D}) &= p(\mathbf{W}^1|\mathcal{D})\cdot p(\mathbf{W}^2|\mathbf{W}^1,\mathcal{D})\cdot \ldots \cdot p(\mathbf{W}^L|\{\mathbf{W}^\ell\}_{\ell=1}^{L-1}, \mathcal{D}) \\
    &= \prod_{\ell=1}^Lp(\mathbf{W}^\ell|\{\mathbf{W}^{\ell'}\}_{\ell'=1}^{\ell-1}, \mathcal{D}).
\end{align*}If we assume that data corresponding to the output of any given layer is available, which of course is not the case, then the conditional posterior for the each layer is of the form
\begin{equation*}
    p(\mathbf{W}^\ell|\{\mathbf{W}^{\ell'}\}_{\ell'=1}^{\ell-1}, \mathcal{D}) \propto \prod_{d=1}^{D^\ell}p(\mathbf{w}_d^\ell)\mathcal{N}\left(\mathbf{y}_d^\ell;\psi(\mathbf{F}^{\ell-1})\mathbf{w}_d^\ell, \left[{\boldsymbol{\Lambda}^\ell_d}\right]^{-1} \right)
\end{equation*}where $\mathbf{y}^\ell_d$ represents all training inputs corresponding to the output of neuron $d$ in layer $\ell$, and $\boldsymbol{\Lambda}^\ell$ is the corresponding precision matrix for the same neuron. In the real world, however, we only have data corresponding to the output layer. As a result, the conditional posterior for the final layer can always be computed exactly via Bayesian linear regression \citep{books/lib/RasmussenW06}, but the same is not true for the other layers. Motivated to introduce pseudo-observations at the outputs of earlier layers by this fact, \citeauthor{ober2021global} posit an approximate posterior distribution defined as $q_{\phi}(\mathbf{W}) = \prod_{\ell = 1}^L q_{\phi}(\mathbf{W}^{\ell} | \{\mathbf{W}^{\ell'}\}_{\ell' = 1}^{\ell - 1}, \mathbf{U}^0)$, where
\begin{equation*}
    q_{\phi}(\mathbf{W}^{\ell} | \{\mathbf{W}^{\ell'}\}_{\ell' = 1}^{\ell - 1}, \mathbf{U}_0) \propto \prod_{d = 1}^{D^{\ell}} p(\mathbf{w}^{\ell}_d)  \mathcal{N}\left(\mathbf{v}^{\ell}_d;\psi(\mathbf{U}^{\ell-1})\mathbf{w}^{\ell}_d,\left[{\boldsymbol{\Lambda}^{\ell}_d}\right]^{-1}\right)
\end{equation*}
such that the full approximate posterior is factorised as approximate conditional posteriors over weights for each layer, mirroring the chain rule decomposition of the exact posterior. This built-in structure allows for network weight correlations to be modelled across layers. $\mathbf{U}_0 \in \mathbb{R}^{M \times D}$ represents $M$ \textit{global inducing locations}, with $\{\mathbf{U}^{\ell}\}_{\ell = 1}^L$ defined as
\begin{equation*}
    \mathbf{U}^1 = \mathbf{U}^0 \mathbf{W}_1, \quad \mathbf{U}^{\ell} = \psi(\mathbf{U}^{\ell - 1})\mathbf{W}^{\ell} \quad \text{for }\ell\in\{2, \ldots, L\}.
\end{equation*}
$\mathbf{v}^{\ell} \in \mathbb{R}^{M}$ and $\boldsymbol{\Lambda}^{\ell}_d \in \mathbb{R}^{M \times M}$ denote the means and precisions of \textit{pseudo likelihoods} at each layer corresponding to the M inducing locations, which form the conditional posteriors when multiplied by the prior. The inducing point covariances can be assumed to be the same for each neuron within a layer such that $\boldsymbol{\Lambda}^\ell_d = \boldsymbol{\Lambda}^\ell$. The variational parameters $\phi = \{\mathbf{U}_0, \{\{\mathbf{v}^{\ell}_d, \boldsymbol{\Lambda}^{\ell}_d\}_{d = 1}^{D^{\ell}}\}_{\ell = 1}^L\}$ are then trained by maximising the ELBO as above. Since the means of the pseudo likelihoods represent pseudo observations, we refer to this variational approximate posterior as the Pseudo-Observation Variational Inference BNN (POVI-BNN).

\begin{figure}[h]
    \centering
    \includegraphics[width=0.7\textwidth]{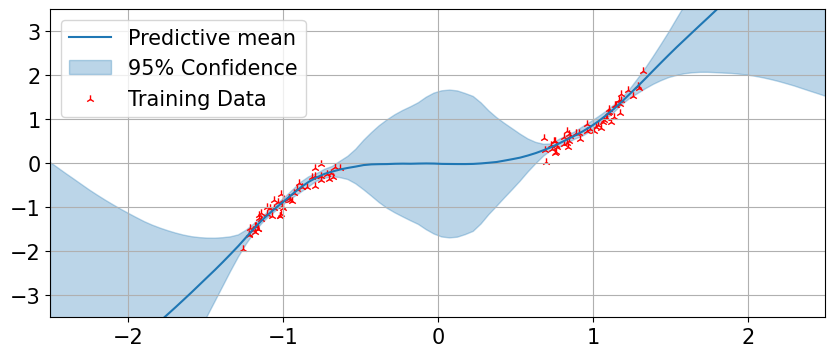}
    \caption{POVI-BNN predictive distribution on a toy regression example, the noisy cubic dataset with a central gap as used in \cite{ober2021global}. The predictive distribution was approximated by moment-matching a Gaussian to the sample mean and variance over 100 BNN prediction samples.}
    \label{fig:povi-bnn}
\end{figure}

\cite{ober2021global} compare their POVI-BNN to both an MFVI BNN and an HMC BNN \citep{NIPS1992_f29c21d4}, and demonstrate state-of-the-art performance in the POVI-BNN across a number of regression and classification tasks. Furthermore, \cite{buibiases} shows that the estimate of the marginal likelihood provided by the ELBO of the POVI-BNN is very close to the true marginal likelihood, and, since the difference is the KL divergence between the true and approximate posteriors, that the approximation must be a very good one. Indeed the superior predictive performance of the POVI-BNN over the MFVI-BNN can be seen in Figure \ref{fig:povi-bnn}, a reproduction of one of the primary figures from \cite{ober2021global} that demonstrates the impressive degree to which the POVI-BNN models uncertainty between clusters of datapoints---a behaviour that is rarely seen in models incapable of modelling weight dependencies across layers.

\subsection{Amortised Inference}

It is often the case that we, as humans, need to take time to consider a problem in detail before reaching a conclusion, but that if we encounter a similar problem, we can then reach a similar conclusion very quickly. \cite{Gershman2014AmortizedII} argue that this is because the brain does not perform inference from scratch for every problem encountered, but rather it uses sub-computations from one inference problem on new but related problems. They describe the process of spreading out the computation for a particular problem over multiple problems in this way as \textit{amortisation}.

In the context of probabilistic inference, amortisation is the process of resorting to a secondary inference procedure to find a mapping from observations $\mathcal{D}$ to posterior distributions $p(\theta|\mathcal{D})$. It is generally used when there is enough data that the secondary models can be trained to produce inferences that are very close to the \say{correct} inferences, but with the great advantage that the inference is made at test time---that there are \textit{significant} speed savings. By far the most prevalent application of amortised inference is within a variational setting; amortised variational inference (AVI). In AVI, the goal of the secondary model is to provide an accurate parameterisation of the optimal variational approximate posterior distribution (for a given family of approximate posteriors). Any parameters of the secondary model then become variational parameters, and they are trained by maximising the ELBO as usual. The most flexible, and common as a result, class of secondary inference model is a simple neural architecture, typically a fully-connected Multilayer Perceptron (MLP). From this point onwards it is assumed that the secondary model of choice is an MLP, and as a result we refer to secondary models as secondary inference networks.

\subsubsection{Variational Auto-Encoders}

Perhaps the most significant use of AVI has been for Variational Auto-Encoders (VAE's) \citep{Kingma2014}. A VAE is a probabilistic generative model that is used to generate new samples from the distribution from which the input data is drawn. A VAE consists of two main components. The first is a secondary inference network that is referred to as the \textit{recognition model} or \textit{probabilistic encoder}, and its purpose is to learn to approximate the posterior distribution over some latent low-dimensional representation $\mathbf{z}_n$ of a datapoint $\mathbf{x}_n$, $q_\phi(\mathbf{z}_n)\simeq p(\mathbf{z}_n|\mathbf{x}_n)$. Samples are then taken from this approximate posterior and passed through the second component, the \textit{probabilistic decoder}, which parameterises a likelihood $p_\theta(\mathbf{x}_n|\mathbf{z}_n)$ for a given representation $\mathbf{z}_n$. Since the goal of a VAE is to learn the distribution of the dataset $\mathcal{D}$ so that samples may be drawn, a natural training objective is to maximise $p(\mathcal{D})$. Since this is typically intractable, we can instead optimise a lower bound for it---the ELBO---and learn the parameters of the secondary inference network and the decoder $\{\phi, \theta\}$ jointly. Note that to use the form of the ELBO defined above, we exchange $\mathbf{z}$ and $\boldsymbol{\theta}$

Although \citeauthor{Kingma2014} do not explicitly view the operation of a VAE from an amortised perspective, it certainly can be. The task of performing inference over the low dimensional representation $\mathbf{z}_n$ of a given high dimensional datapoint $\mathbf{x}_n$ is not performed in entirety for each datapoint. Instead, the recognition model amortises the inference, leveraging the fact that much of the inference computation for each datapoint-representation pair is the same, within a given dataset.

\subsection{Neural Processes}
Before proceeding with the narrative of amortising inference in BNNs, at this point we introduce a collection of models that will serve as a benchmark in probabilistic meta-learning: the Neural Process Family (NPF) \citep{dubois2020npf}.

\subsubsection{The Neural Process Family}

Following the usual Bayesian machine learning procedure, one must specify both a prior distribution over model parameters and a model for data (the likelihood). One then performs inference to obtain a posterior distribution over model parameters, which, after multiplication by the likelihood of a test datapoint and marginalisation of the parameters, gives a posterior predictive distribution. It is a somewhat roundabout procedure if we are only really interested in the posterior predictive distribution. The Neural Process Family \citep{garnelo2018conditional, garnelo2018neural} is a collection of probabilistic metamodels that meta-learn to make posterior predictions directly, effectively cutting out the oftentimes problematic \say{middle man} of performing inference. Further, they combine the practical flexibility and predictive power of neural networks with the desirable uncertainty properties of stochastic process models, all in a meta-learning regime \citep{dubois2020npf} in which predictions for unseen datasets are made in a single forward pass. 

At the highest level, an NP is a model consisting of an encoder $\text{Enc}(\cdot)$ and a decoder $\text{Dec}(\cdot)$, both of which comprise some neural architecture. The encoder generates a representation of a dataset in a single forward pass, which is then used in some way by the decoder along with a test location of interest to generate a posterior prediction. To avoid ambiguity later on, we refer to datapoints with known labels within a given dataset as \textit{context} datapoints, and test datapoints as \textit{target} datapoints. Note that for datasets used to train a meta-model (training datasets), the labels are known for both context and target points, but for an unseen dataset (test dataset) we only have access to context input-output pairs.

\subsubsection{The Conditional Neural Process Family}

The NP family can be split into two subfamilies; the conditional NP family (CNPF) and the latent NP family (LNPF), but for a particular type of NP there is in general both a conditional and a latent variant. In a member of the CNPF, the encoder maps from the context dataset $\mathcal{D}_\mathcal{C}$ to a deterministic representation $R$. This is done in a permutation invariant manner so that the dataset is indeed treated as a \textit{set}. The decoder maps from the representation $R$ directly to a conditional (on the representation) posterior predictive distribution $p_\theta(\mathbf{y}_\mathcal{T}|\mathbf{x}_\mathcal{T},R) = p_\theta(\mathbf{y}_\mathcal{T}|\mathbf{x}_\mathcal{T},\mathcal{D}_\mathcal{C})$. Note that the decoder is parameterised by $\theta$. Since predictive distributions for different target locations $\mathbf{x}_\mathcal{T}$ need to be consistent with each other, the more general constraint that guarantees such consistence
\begin{equation*}
    p_\theta(\mathbf{y}_\mathcal{T}|\mathbf{x}_\mathcal{T},R) = \prod_{t=1}^{|\mathcal{T}|}p_\theta\left(\mathbf{y}_\mathcal{T}^{(t)}|\mathbf{x}_\mathcal{T}^{(t)},R\right), \quad \text{where }R = \text{Enc}(\mathcal{D}_\mathcal{C})
\end{equation*}is enforced, and so for regression contexts the predictive likelihood is typically chosen to be Gaussian---a logical but somewhat restrictive move. Members of the CNPF have the advantage that inference is always tractable, partially because a forward pass is entirely deterministic. However, since the conditional predictive distributions for a collection of target points are independent of each other, it is not possible to draw coherent functional samples from the distribution in the way it is for, say, a Gaussian Process (GP) \citep{books/lib/RasmussenW06}.

\subsubsection{The Latent Neural Process Family}

By contrast, members of the LNPF encode context datasets into a stochastic latent variable $\mathbf{z} \sim p_{\theta'}(\mathbf{z}|R)$, where $\theta'$ denotes the parameters of the encoder. In practice, however, the encoder is used to generate a deterministic representation $R$ as before, which is then used to parameterise the distribution over $\mathbf{z}$---again this distribution is typically chosen to be Gaussian. The decoder then maps from the latent variable $\mathbf{z}$ to the $\mathbf{z}$-dependent posterior predictive distribution $p_\theta(\mathbf{y}_\mathcal{T}|\mathbf{x}_\mathcal{T},\mathbf{z})$. The $\mathcal{D}_\mathcal{C}$-dependent posterior predictive distribution that we are actually interested in is then given by
\begin{equation*}
    p\left(\mathbf{y}_\mathcal{T}|\mathbf{x}_\mathcal{T},\mathcal{D}_\mathcal{C}\right) = \int p_{\theta'}\left(\mathbf{z}|R\right)p_\theta\left(\mathbf{y}_\mathcal{T}|\mathbf{x}_\mathcal{T},\mathbf{z}\right)\mathrm{d}\mathbf{z}.
\end{equation*}If the distribution over the latent variable is indeed Gaussian, then the overall posterior predictive distribution can be viewed as an infinite mixture of Gaussians \citep{bishop2007}, and so can take \textit{any} form. Furthermore, dependencies can be maintained across target locations $\mathbf{x}_\mathcal{T}^{(t)}$ and as a result we can draw coherent samples from the posterior predictive distribution. Unfortunately, computing the posterior predictive distribution is intractable for any member of the LNPF with practically useful modelling power, and so we must resort to approximations.

\subsubsection{Training in the Neural Process Family}

For members of the CNPF, training is straightforward since the posterior predictive likelihood is tractable, and so we use the neural process maximum likelihood (NPML) objective:
\begin{align*}
    \mathcal{L}_{NPML} &= p_\theta(\mathbf{y}_\mathcal{T}|\mathbf{x}_\mathcal{T}, \mathcal{D}_\mathcal{C}) \\
    &= p_\theta(\mathbf{y}_\mathcal{T}|\mathbf{x}_\mathcal{T}, R),\quad R = \text{Enc}(\mathcal{D}_\mathcal{C}).
\end{align*}For members of the LNPF, however, the predictive likelihood is intractable and so we are left with two options. The first is to approximate the predictive likelihood by applying Monte Carlo integration
\begin{align*}
    \mathcal{L}_{NPML} &= p_\theta(\mathbf{y}_\mathcal{T}|\mathbf{x}_\mathcal{T}, \mathcal{D}_\mathcal{C}) \\
    &\simeq \frac{1}{M}\sum_{m=1}^Mp_\theta\left(\mathbf{y}_\mathcal{T}|\mathbf{x}_\mathcal{T},\mathbf{z}^{(m)}\right), \quad \mathbf{z}^{(m)} \sim p_{\theta'}\left(\mathbf{z}|R\right).
\end{align*}and then proceed as before. The second option is to interpret the distribution over the latent variable $p_{\theta'}(\mathbf{z}|R)$ as an approximation to the posterior distribution over $\mathbf{z}$ given the both the context and target datasets $\mathcal{D}_\mathcal{C}, \mathcal{D}_\mathcal{T}$. Unfortunately, $p(\mathbf{z}|\mathcal{D}_\mathcal{C},\mathcal{D}_\mathcal{T})$ is intractable and so we replace it with the posterior over $\mathbf{z}$ given all the datapoints as if they were a single dataset $\mathcal{D}_\mathcal{C}\cup\mathcal{D}_\mathcal{T}$, which is simply computed by passing both the context and target datasets through the encoder:
\begin{align*}
    p_{\theta'}(\mathbf{z}|R) &= p_{\theta'}(\mathbf{z}|\mathcal{D}_\mathcal{C}) \\
    &\simeq p(\mathbf{z}|\mathcal{D}_\mathcal{C},\mathcal{D}_\mathcal{T}) \\
    &\simeq p_{\theta'}(\mathbf{z}|\mathcal{D}_\mathcal{C}\cup\mathcal{D}_\mathcal{T})
\end{align*}and then carry out VI. The objective used in such a procedure is known as the neural process variational inference (NPVI) objective and is given by
\begin{equation*}
    \mathcal{L}_{NPVI} = \mathbb{E}_{p_{\theta'}(\mathbf{z}|\mathcal{D}_\mathcal{C}\cup\mathcal{D}_\mathcal{T})}\left[\log p_\theta(\mathbf{y}_\mathcal{T}|\mathbf{x}_\mathcal{T},\mathbf{z})\right] - \text{KL}\left[p_{\theta'}(\mathbf{z}|\mathcal{D}_\mathcal{C}\cup\mathcal{D}_\mathcal{T})\|p_{\theta'}(\mathbf{z}|\mathcal{D}_\mathcal{C})\right].
\end{equation*}Note that the expected log likelihood is generally intractable and so we usually approximate it using Monte Carlo integration---similar to when we compute the ELBO in BNNs. The left-hand term in the NPVI objective encourages a good fit to the \textit{full} dataset, while the second term encourages similar behaviour when just the context dataset is observed.

The question of which objective to use when training members of the LNPF remains an open one. \citeauthor{foong2020metalearning} argue that the NPVI objective potentially over-prioritises the pursuit of consistent posterior approximations, and that the NPML objective should be used instead since obtaining high quality posterior predictive distributions is what we are ultimately interested in.

\subsubsection{Convolutional Conditional Neural Processes}
A particular member of the NPF that is worth introducing is the Convolutional Conditional Neural Process (ConvCNP) \citep{gordon2020convolutional}, since we will use it to represent the state-of-the-art (SOTA) across a number of probabilistic meta-learning problems.

For many meta-learning applications, we would like predictions to be consistent regardless of the absolute position of inputs. An example of this is in an image completion setting; if there are pixels in an incomplete image that make up part of, say, a face, we would like our model to exhibit the same ability to recognise and reconstruct a face regardless of where in the image the pixels are, and crucially the reconstructions should be shifted in the same way as the inputs. Such behaviour is achieved when a model is \textit{translation equivariant}. Translational equivariance was the key motivator for the invention of Convolutional Neural Networks (CNNs) (\citealt{10.1162/neco.1989.1.4.541}, \citealt{Fukushima1980NeocognitronAS})---one of the better-known success stories of deep learning---and so this motivates the use of convolutional architectures in neural processes in order to achieve translational equivariance. 

Unfortunately, CNNs operate on signals that lie on discrete domains, but we would like to query our translationally equivariant representation arbitarily. As such, the problem is not as simple as exchanging standard neural architectures for convolutional ones. To obtain flexible and translationally equivariant representations, we are motivated to map our dataset to a continuous, \textit{functional}, representation. To solve this issue, \citeauthor{gordon2020convolutional} introduce a type of convolution that operates on sets, the \textit{SetConv}:
\begin{equation*}
    \text{SetConv}\left(x, \{(x^{(c)}, y^{(c)}\}_{c=1}^{|\mathcal{D}_\mathcal{C}|}\right) = \sum_{c=1}^{|\mathcal{D}_\mathcal{C}|}
    \begin{pmatrix}
    1 \\
    y^{(c)}
    \end{pmatrix}
    w_\theta\left(x-x^{(c)}\right).
\end{equation*}The additional dimension with the value 1 is included to ensure that context points with output value 0 and a missing point are distinguishable, and is referred to as a \textit{density channel}. $w_\theta(\cdot)$ is a distance function, and is typically chosen to be a Euclidean radial basis function with a learnable scale parameter $\sigma_l$
\begin{equation*}
    w_\theta(\cdot) = e^{-\frac{\|\cdot\|_2^2}{\sigma_l^2}}.
\end{equation*}

The SetConv maps from arbitrary sets of datapoints to continuous functions, but these functions can be evaluated at regular intervals in order to produce a griddled representation of a dataset which can then be passed through a CNN. The output of the CNN can then be passed through another SetConv in order to produce a functional representation $R$. This representation can then be evaluated at the target locations to produce a sequence of target-dependent functional representation evaluations (vectors), each of which are passed through the decoder to obtain the conditional posterior predictive distribution at the corresponding target location. 

\begin{figure}[h]
    \centering
    \includegraphics[width=0.7\textwidth]{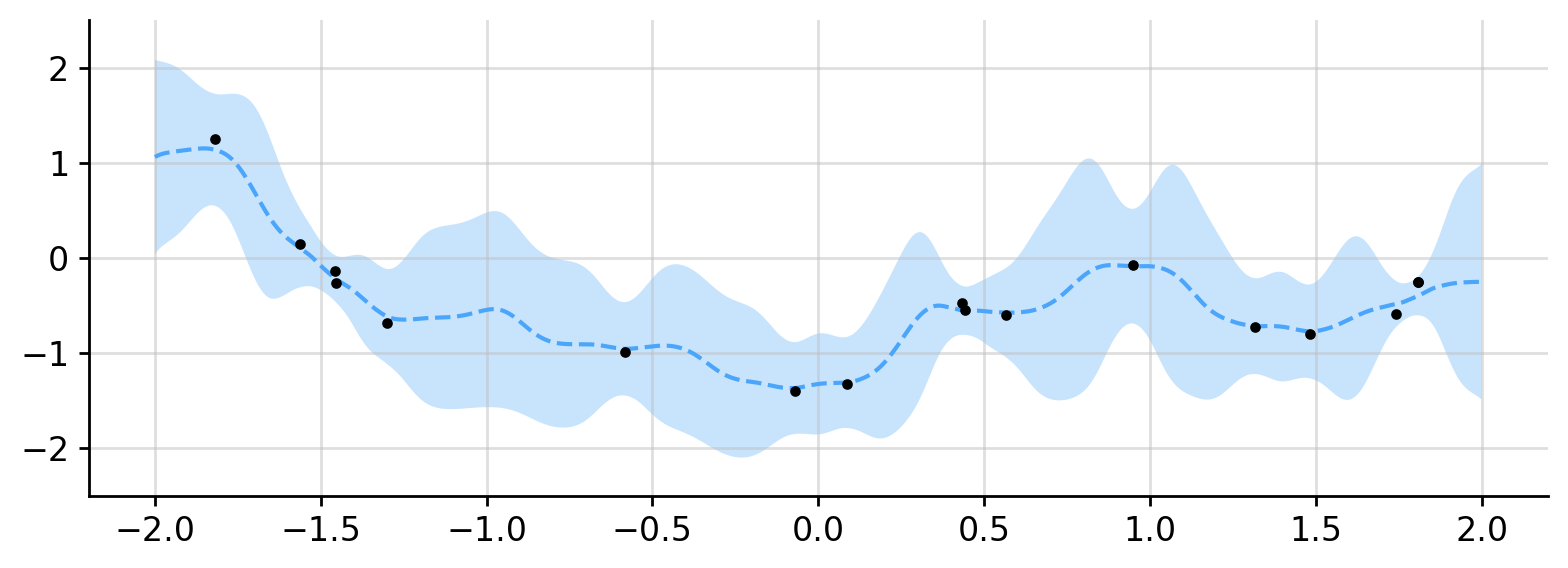}
    \includegraphics[width=0.7\textwidth]{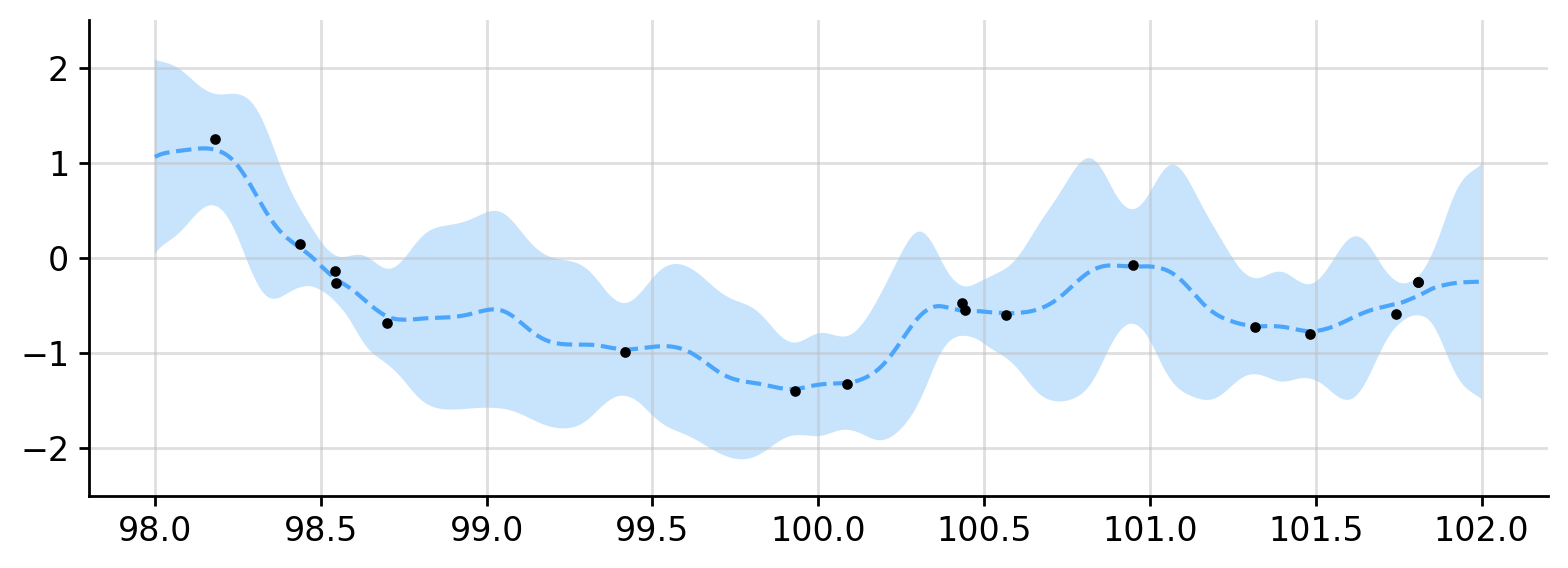}
    \caption{ConvCNP predictive distributions for a toy regression task generated from a GP prior sample with Laplacian covariance function. The task in the second plot is the task from the first plot shifted 100 units to the right. The ConvCNP was trained on a meta-dataset of similarly produced tasks that were all centered at 0. The dashed line indicates the predictive mean, the shaded region indicates the 95\% confidence zone, and the black dots are context points.}
    \label{fig:convcnp}
\end{figure}

Such a framework specifies a ConvCNP, or more specifically, an \textit{off-the-grid} ConvCNP since the data on which it operates does not have to be griddled. Note that the distinction between encoder and decoder is somewhat arbitrary, and the output of the first SetConv can be used as the representation instead \citep{dubois2020npf}. In cases where the dataset is already on a grid, the SetConvs are not necessary; the data are ready for standard convolution immediately after appending the density channel---a ConvCNP variant termed the \textit{on-the-grid} ConvCNP. Note that datasets are represented by a point in a reproducing kernel Hilbert space (RKHS). Since function spaces such as Hilbert spaces can be viewed as an infinite dimensional vector space, this is another advantage that ConvCNPs have over existing NPs which only embed datasets into finite dimensional vector representations. Moreover, there are many positive implications of embedding the datasets into both a Hilbert space and an RKHS in particular \citep{gordon2020convolutional}, as well as the interesting fact that similarities with GPs---which also operate over RKHS's---can be drawn. Figure \ref{fig:convcnp} demonstrates the translational equivariance of ConvCNPs.

\section{Amortising Inference in BNNs}

In some cases, amortisation can be performed in a \textit{per-datapoint} manner; each datapoint is passed through a secondary inference network and the output is combined in some permutation invariant manner with that of the other datapoints to provide the amortised variational approximate posterior. In this case the amortised model is particularly suitable to be used in a meta-learning context since it is easy to handle datasets of variable sizes\footnote{Note that if we were to perform per-dataset amortised inference, we would need a way to embed datasets into a vector that is of the correct, fixed, size to be used to parameterise the approximate posterior distribution. In such a setting the secondary inference network would have to be something that can handle variable-length sequences, for example an RNN.}. The implication of this is that we can easily extend a variational model to a variational meta-model through amortisation if the variational approximate posterior decomposes into a product of per-datapoint factors. The model can then be trained by sampling datasets from the meta-dataset and computing the ELBO for each dataset, carrying out a single update step of gradient descent after exposure to the minibatch size number of datasets---in other words the usual gradient descent scheme except datasets within the meta-dataset are treated like datapoints within a dataset.

\subsection{Amortised MFVI in BNNs}
The form of the MFVI approximate posterior presented above does not decompose into a product of per-datapoint factors. However, exploiting the fact that products of Gaussian distributions yield (unnormalised) Gaussian distributions, we can alter the form of the approximate posterior to 
\begin{equation*}
    q(\mathbf{W}) \propto p(\mathbf{W})\prod_{n=1}^N\mathcal{N}\left(\mathbf{W};\boldsymbol{\mu}_n,\boldsymbol{\Sigma}_n\right)
\end{equation*}where $\boldsymbol{\mu}_n$ and $\boldsymbol{\Sigma}_n$ are the mean vector and diagonal covariance matrix that correspond to an approximate likelihood for a single datapoint. As long as the prior $p(\mathbf{W})$ is mean-field Gaussian, which is typically the case, then the posterior is also mean-field Gaussian:
\begin{equation*}
    q(\mathbf{W}) \propto \prod_{i=1}^{|\mathbf{W}|}\mathcal{N}\left(w_i;0,\sigma_p^2\right)\prod_{n=1}^N\mathcal{N}\left(w_i;\mu_{n,i},\left[{\Sigma_n}\right]_{i,i}^2\right)
\end{equation*}This model is then amortised by passing each datapoint pair $(\mathbf{x}_n, \mathbf{y}_n)$ through a secondary inference network to obtain the corresponding mean and covariance matrix diagonal vector\footnote{Note that in practice we use the secondary inference networks to obtain the \textit{logarithm} of variance parameters instead in order to circumvent non-negativity constraints.\label{fn:logvar}}:
\begin{equation*}
    \begin{pmatrix}
    \boldsymbol{\mu}_n \\
    \boldsymbol{\sigma}_{n}
    \end{pmatrix} = g_\phi(\mathbf{x}_n,\mathbf{y}_n)
\end{equation*}where $\boldsymbol{\mu}_n,\text{ }\boldsymbol{\sigma}_{n}\in\mathbb{R}^{|\mathbf{W}|}$ and $\boldsymbol{\sigma}_n = \text{Diag}\left(\boldsymbol{\Sigma}_n\right)$. We refer to this model as the Amortised MFVI-BNN (AMFVI-BNN).

\subsection{Amortised POVI in BNNs}
This section covers our core contribution, the Amortised POVI-BNN (APOVI-BNN), and we begin by building upon the machinery introduced by \citeauthor{ober2021global}. 

Similarly to that of the MFVI-BNN, the approximate posterior for the POVI-BNN does not decompose across datapoints. However, this can be resolved by setting the inducing points to be the available datapoints:
\begin{equation*}
    \mathbf{U}_0 = \mathbf{X}
\end{equation*}and by assuming the pseudo-likelihood covariance matrices to be diagonal. The layerwise conditional approximate posteriors can then be written as
\begin{equation*}
    q_\phi\left(\mathbf{W}^\ell|\{\mathbf{W}^{\ell'}\}_{\ell'=1}^{\ell-1},\mathcal{D}\right) \propto \prod_{d=1}^{D^\ell}p(\mathbf{w}_d^\ell)\prod_{n=1}^N\mathcal{N}\left(V_{n,d}^\ell;x^{\ell}_{n,d},{\Sigma_{n,d}^\ell}^2\right)
\end{equation*}where
\begin{equation*}
    \mathbf{x}_n^0 = \mathbf{x}_n,\quad \mathbf{x}_n^1 = \mathbf{W}^1\mathbf{x}_n, \quad \mathbf{x}_n^\ell = \mathbf{W}^\ell\psi\left(\mathbf{x}_n^{\ell-1}\right)\text{ for }\ell\in\{2,...,L\}.
\end{equation*}with $x^{\ell}_{n,d}$ denoting the $d$-th element of $\mathbf{x}_{n}^{\ell}\in\mathbb{R}^{D^{\ell}}$ and with $\mathbf{V}^\ell,\text{ }\boldsymbol{\Sigma}^\ell\in\mathbb{R}^{N\times D^\ell}$. If each BNN layer is endowed with a secondary inference network such that layer $\ell$ has corresponding inference network $g_\phi^\ell(\cdot)$, then amortisation\footref{fn:logvar} is carried out as 
\begin{equation*}
    \begin{pmatrix}
    \mathbf{v}_n^\ell \\
    \boldsymbol{\sigma}_n^\ell
    \end{pmatrix} = g_\phi^\ell\left(\mathbf{x}_n,\mathbf{y}_n\right)
\end{equation*}where $\mathbf{v}_n^\ell,\text{ }\boldsymbol{\sigma}_n^\ell\in\mathbb{R}^{D^\ell}$ are being used instead of the clumsier notation $\left[\mathbf{V}_{n,:}^\ell\right]^T,\text{ }\left[\boldsymbol{\Sigma}_{n,:}^\ell\right]^T$. For the final layer we have access to the output data $\mathbf{y}$ and as such the secondary inference network is only used to predict variances. For a mean-field Gaussian prior, and following standard results for posterior distributions that are proportional to products of Gaussians, the approximate posterior for the weights of layer $\ell$ conditioned on those of previous layers is given by
\begin{equation*}
    q_\phi\left(\mathbf{W}^\ell|\{\mathbf{W}^{\ell'}\}_{\ell'=1}^{\ell-1},\mathcal{D}\right) = \prod_{d=1}^{D^\ell}\mathcal{N}\left(\mathbf{w}_d^\ell;\boldsymbol{\Sigma}^\ell_\mathbf{w}\psi\left(\mathbf{X}^{\ell-1}\right)^T\boldsymbol{\Lambda}^{\ell}_d\left[\mathbf{V}^{\ell}\right]_{:,d},\boldsymbol{\Sigma}^\ell_\mathbf{w}\right)
\end{equation*}
where
\begin{equation*}
    \boldsymbol{\Sigma}^\ell_\mathbf{w} = \left(\sigma_p^{-2}\mathbf{I}_{D^\ell} + \psi\left(\mathbf{X}^{\ell-1}\right)^T\boldsymbol{\Lambda}_d^\ell\psi\left(\mathbf{X}^{\ell-1}\right)\right)^{-1}
\end{equation*}
and $\boldsymbol{\Lambda}^\ell_d$ is the $N\times N$ diagonal precision matrix that corresponds to pseudo outputs of neuron $d$ in layer $\ell$ and which is obtained from the secondary inference network variance outputs by
\begin{equation*}
    \left[\boldsymbol{\Lambda}^\ell_d\right]_{n,n} = \left({\Sigma_{n,d}^\ell}\right)^{-2}.
\end{equation*}Similarly, note that $\left[\mathbf{V}^\ell\right]_{:,d}\in\mathbb{R}^{N}$ is a vector of pseudo outputs corresponding to the same neuron. Finally, note that $\mathbf{X}^{\ell}\in\mathbb{R}^{N\times D^\ell}$ is a stack of the transposes of the $\mathbf{x}_n^\ell$ vectors for $n\in\{0,...,\ell\}$.

It is important to note that this framework requires the entire dataset $\mathcal{D}$ to be propagated through the network in order to compute $\boldsymbol{\Sigma}_w^\ell$ for each layer $\ell$. This does not pose a problem for smaller datasets, but since there is no obvious way to perform mini-batching \textit{within a dataset} this method is not expected to be of practical use for very large datasets, since we would need to store all datapoints in memory. It is for this reason that we do not extend \citeauthor{ober2021global}'s convolutional POVI-BNN to the amortised setting as well; CNN's are of most use on high dimensional data within very large datasets (e.g. images), and our method has too high memory requirements for such applications.

\subsection{The APOVI-BNN as a Neural Process} \label{sec:apovi-np}
To summarise the similarity between the operation of the APOVI-BNN and that of a member of the LNPF, we can break a forward pass of the model down into four distinct steps that are described in terms of  equivalent steps in a member of the LNPF:
\begin{enumerate}
    \item Using the (neural) secondary inference networks, generate a representation for each datapoint. This representation parameterises the per-datapoint likelihoods in the APOVI-BNN.
    \item Aggregate these representations in a permutation-invariant manner to generate a representation for the dataset as a whole. This is done by computing the product of pseudo-likelihoods, and is done implicitly during the following step.
    \item Use the representation of the dataset to parameterise a distribution over a latent variable. This is done when the pseudo likelihoods are used to find the approximate posterior distribution over model weights.
    \item Sample from the latent variable to parameterise a mapping from inputs to predictive samples. In the APOVI-BNN we sample the model weights to compute posterior prediction samples at test locations.
\end{enumerate}

The similarities are clear, and there is only one major difference between the two models---in the APOVI-BNN the parameters of the decoder (the primary network) are task-specific, but in a member of the LNPF these parameters are shared. Furthermore, the parameters of the decoder are also the latent variable. 

Nevertheless, the similarities are strong enough to motivate training the APOVI-BNN as if it were a member of the LNPF. If we use the NPVI objective, modelling power is exerted on ensuring consistency between the approximate and true posterior distributions. Under such a regime, the APOVI-BNN may in theory reap all the usual benefits of the (approximately) Bayesian approach; namely data efficiency and accurate uncertainty estimates. By contrast, if we use the NPML objective then the secondary inference network parameters are no longer variational---they are simply part of the model. As such, the NPML objective is likely to encourage overfitting in the model. Despite this, it could still lead to superior predictions. We expect both NP objectives to yield better performance than the regular ELBO since they encourage the model to predict well on target points that are not in the context dataset.

\section{Related Work}

\textbf{Latent Neural Processes.} The APOVI-BNN has distinct similarities with members of the LNPF as discussed. \cite{volpp2021bayesian} introduce the idea of \textit{Bayesian aggregation} for use in LNPF-based models, in which an approximate posterior over the latent representation of a context dataset is formed from the product of approximate likelihoods of datapoint representations and the prior. Viewing the model weights in the APOVI-BNN as \citeauthor{volpp2021bayesian}'s latent variable, the resemblance is clear, but it is important to note the difference in form between their latent variable and the BNN weights.

\textbf{Amortised Variational Inference.} There are strong similarities between our use of secondary inference networks within a VI framework and VAEs \citep{Kingma2014}. Since their introduction, however, applications of AVI have been extended to other probabilistic models. Examples include applications to graphical models (\citealt{NIPS2016_7d6044e9}, \citealt{lin2018variational}), as well as to VAE models that use GP priors to allow for dependencies between datapoint dimensions to be modelled (\citealt{ashman2020sparse}, \citealt{jazbec2021scalable}). AVI has also seen applications in causal inference (\citealt{pawlowski2020deep} \citealt{ashman2023causal}).

\textbf{Meta-Learning in Neural Networks.} A very popular area of research in recent times has been the application of meta-learning to neural networks. Perhaps the most notable example is the Model-Agnostic Meta-Learning (MAML) framework \citep{finn2017modelagnostic}. There have been expansions of MAML which try to find a good initialisation of model parameters \citep{antoniou2019train}, Bayesian variants of the framework \citep{kim2018bayesian}, and extensions which, somewhat like the APOVI-BNN, find task-specific parameters that are conditioned on the dataset such as the Meta-Learning Probabilistic Inference for Prediction (ML-PIP) framework \citep{gordon2019metalearning} and Conditional Neural Adaptive Processes (CNAPs) \citep{requeima2020fast}. The key difference between these frameworks and the APOVI-BNN is the fact that, even if they utilise task-specific parameters, they use \textit{very} many shared model parameters and so rely on huge meta-datasets as a result, whereas the APOVI-BNN has no shared model parameters and instead meta-learns inference in a BNN.

\section{Experiments}

In this section we describe three experiments that were performed to examine the capabilities of the APOVI-BNN. Basic experimental details are included in this section, but for the finer details such as model architectures and hyperparameters, please see Appendix \ref{app:exp}.

\subsection{Effect of Amortisation on Approximate Posterior Quality}

We begin evaluation of the APOVI-BNN by investigating the degree to which amortisation degrades the quality of obtained approximate posteriors in comparison to the vanilla POVI-BNN, especially as a function of the size of the meta-dataset $\boldsymbol{\Xi}$ used, $|\boldsymbol{\Xi}|$. We can do this by evaluating the ELBO on test datasets for both the APOVI-BNN and the POVI-BNN. Keeping the model architectures the same, as well as other hyperparameters such as the choice of prior variance $\sigma_p^2$ or observation noise at the output $\sigma_{noise}^2$, the marginal likelihood is identical between the two models. As such, any difference in computed ELBO values is entirely due to differences in the KL divergence between approximate and exact posterior distributions $\text{KL}\left[q_\phi(\mathbf{W})\|p(\mathbf{W}|\mathcal{D})\right]$, similar to the case in \cite{buibiases}.

For this experiment, \textit{all} datasets used are constructed from squared-exponential (SE) covariance GP prior samples, and uncertainty bounds are found by taking averages over 5 different test datasets, which themselves are unchanged between tests. For the APOVI-BNN, each test is repeated 5 times for different randomly generated training meta-datasets, while for the POVI-BNN it is simply trained on each test dataset. The APOVI-BNN is evaluated as both a regular model as well as a meta-model across different sizes of training meta-dataset. When the APOVI-BNN is trained as a regular model on just the test datasets, we refer to the meta-dataset as being of size 0 for notational ease, $|\boldsymbol{\Xi}| = 0$, but it should be stressed that this is the only case in which the APOVI-BNN is exposed to the test datasets during training. Note also that in the this case the APOVI-BNN is trained and tested from scratch for each test dataset as if it were incapable of meta-learning.

\begin{table}[h]
    \centering
    \begin{tabular}{|r||r|r|r|r|r|}
        \hline
        \multicolumn{1}{|l||}{\multirow{2}{*}{POVI-BNN}} & \multicolumn{5}{c|}{APOVI-BNN} \\
        \cline{2-6}
        \multicolumn{1}{|r||}{} & $|\boldsymbol{\Xi}| = 0$ & $|\boldsymbol{\Xi}| = 1$ & $|\boldsymbol{\Xi}| = 2$ & $|\boldsymbol{\Xi}| = 5$ & $|\boldsymbol{\Xi}| = 10$ \\
        \hline \hline
        $3.13\pm0.37$ & $5.38\pm0.59$ & $-0.24\pm6.12$ & $3.22\pm4.05$ & $4.81\pm0.49$ & $4.48\pm0.88$ \\
        \hline
    \end{tabular}
    \caption{Obtained ELBO value for POVI-BNN and APOVI-BNN on test datasets generated from SE covariance GP prior samples.}
    \label{tab:elbo_exp}
\end{table}

We see that when the APOVI-BNN is treated as a regular model, the quality of the posterior approximation obtained surpasses that of the POVI-BNN. When restricted to functioning as a meta-model that only encounters the test-dataset at test time, the APOVI-BNN's performance increases as the number of training datasets increases. It is particularly notable that after exposure to just three training datasets, the APOVI-BNN achieves a significantly better ELBO than the POVI-BNN.

\subsection{One-Dimensional Regressions}

In this experiment we compare the predictive performance of the APOVI-BNN against two other models on one-dimensional artificial datasets. The predictive performance of the APOVI-BNN is compared with that of an AMFVI-BNN and a ConvCNP in two slightly different scenarios; firstly we use test datasets that lie within the distribution of the meta-dataset, and then we use test datasets that are out-of-meta-dataset. 

\begin{figure}[h]
  \centering
  \begin{subfigure}{0.49\textwidth}
  \centering
    \includegraphics[width=\linewidth]{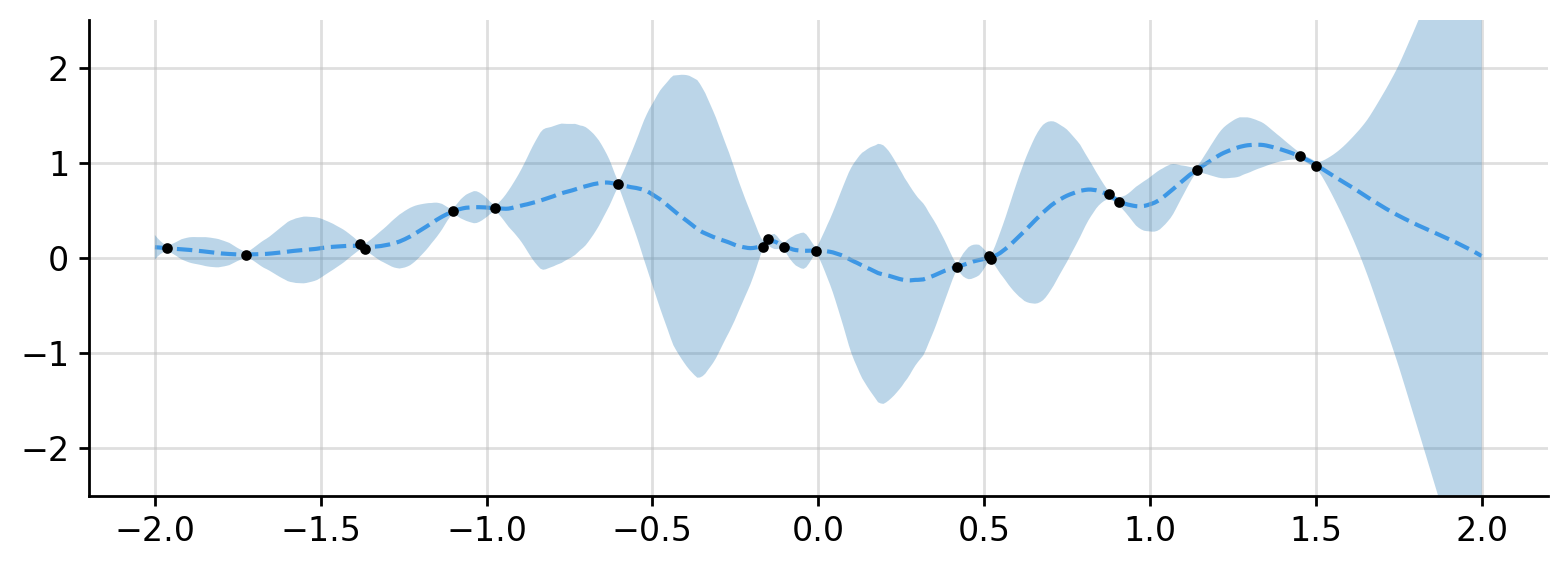}
    \caption{APOVI-BNN, $|\boldsymbol{\Xi}|=1$}
    \label{fig:subfig1}
  \end{subfigure}
  \begin{subfigure}{0.49\textwidth}
  \centering
    \includegraphics[width=\linewidth]{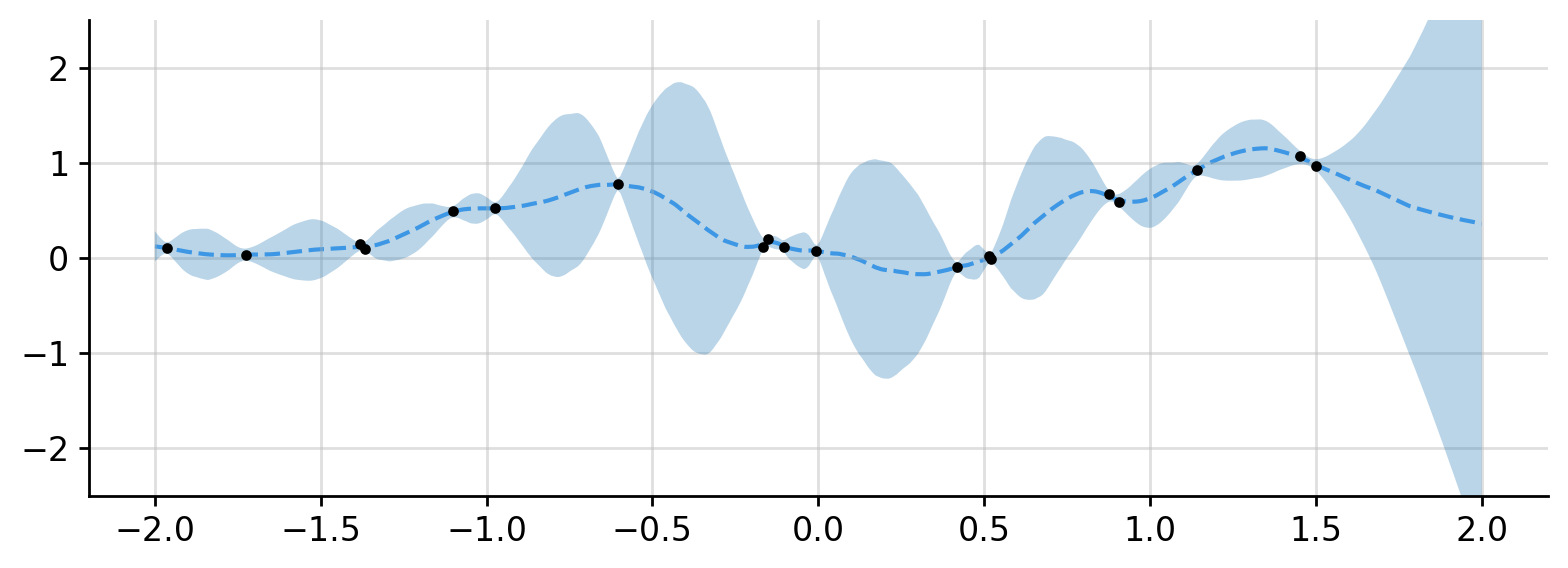}
    \caption{APOVI-BNN, $|\boldsymbol{\Xi}|=100$}
    \label{fig:subfig2}
  \end{subfigure}
  \vspace{0.5cm}
  
  \begin{subfigure}{0.49\textwidth}
  \centering
    \includegraphics[width=\linewidth]{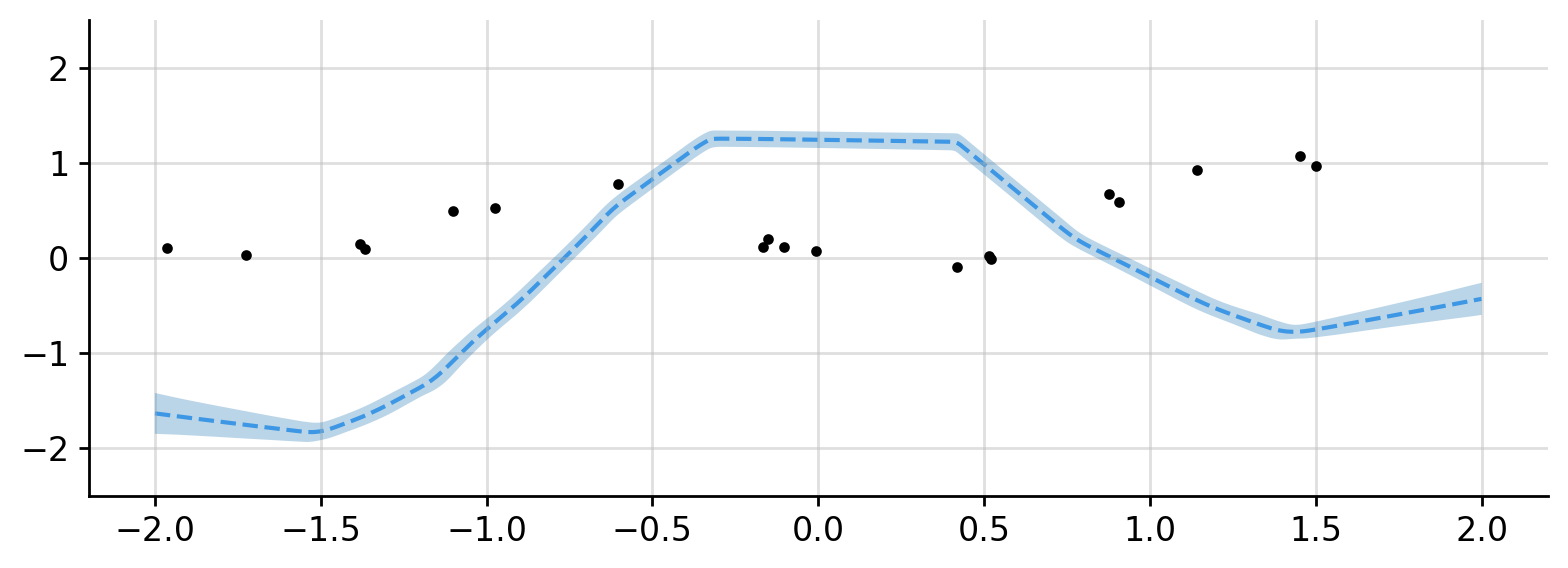}
    \caption{AMFVI-BNN, $|\boldsymbol{\Xi}|=1$}
    \label{fig:subfig3}
  \end{subfigure}
  \begin{subfigure}{0.49\textwidth}
  \centering
    \includegraphics[width=\linewidth]{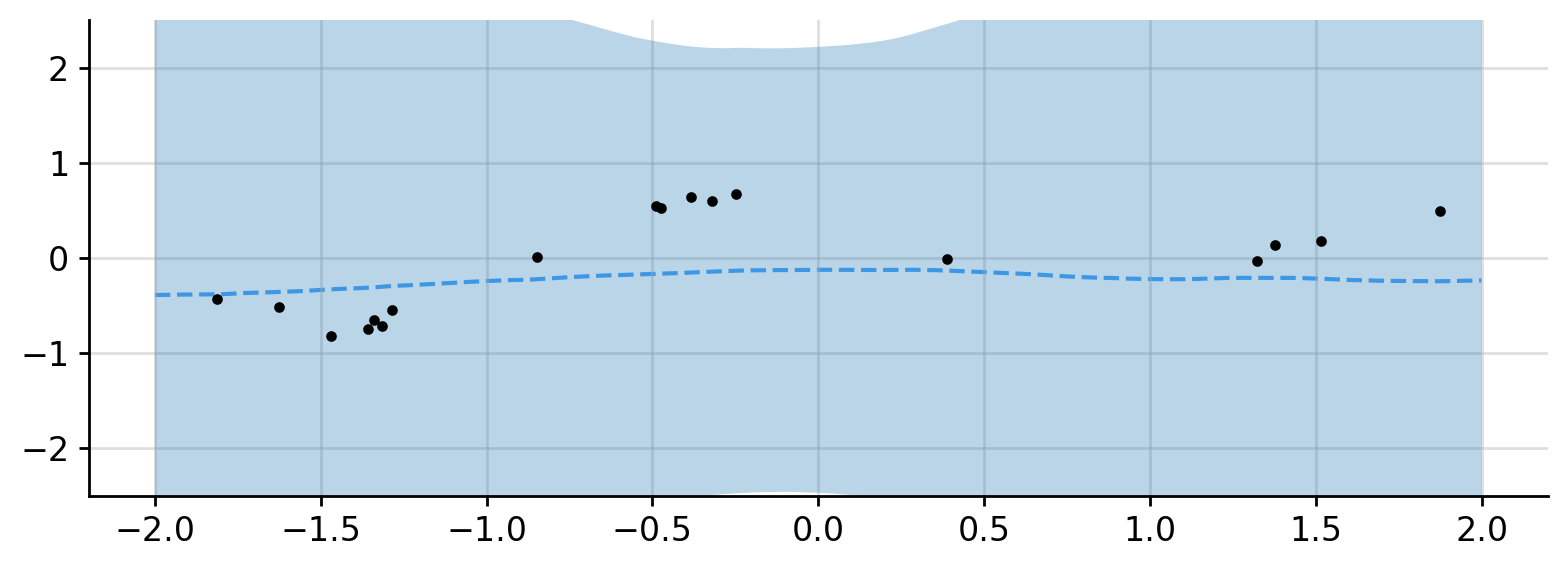}
    \caption{AMFVI-BNN, $|\boldsymbol{\Xi}|=100$}
    \label{fig:subfig4}
  \end{subfigure}
  \vspace{0.5cm}
  
  \begin{subfigure}{0.49\textwidth}
  \centering
    \includegraphics[width=\linewidth]{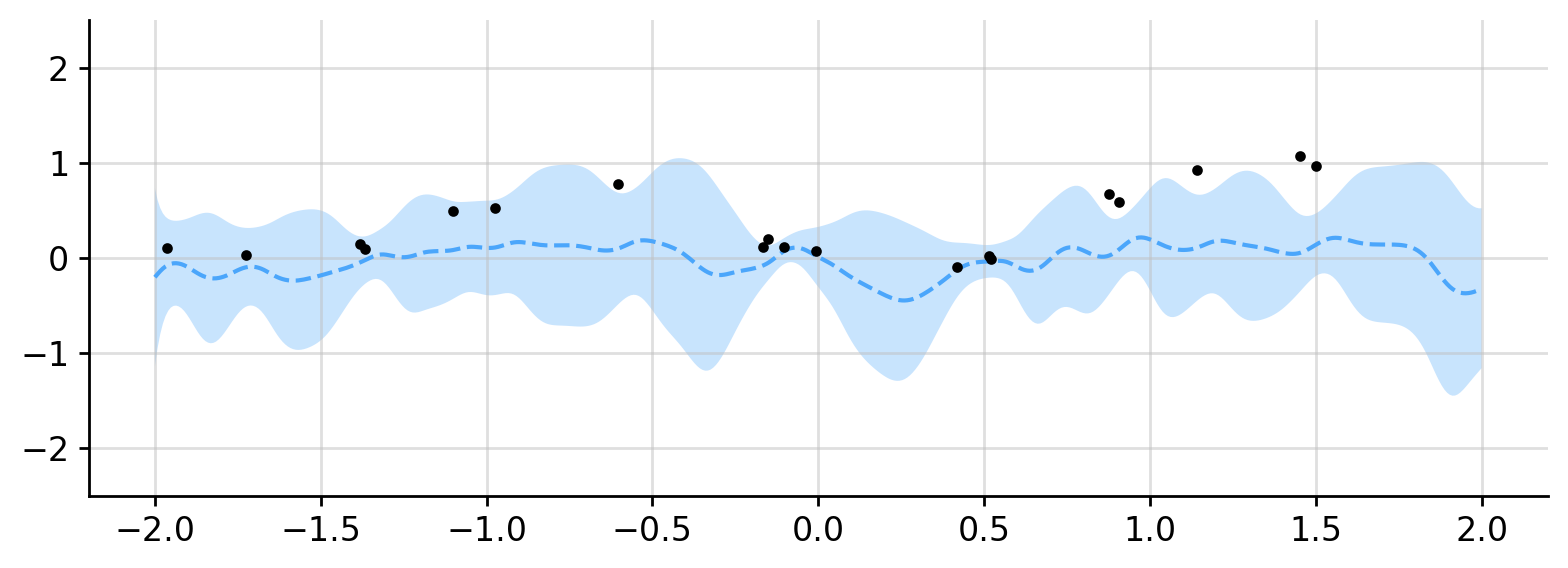}
    \caption{ConvCNP, $|\boldsymbol{\Xi}|=1$}
    \label{fig:subfig5}
  \end{subfigure}
  \begin{subfigure}{0.49\textwidth}
  \centering
    \includegraphics[width=\linewidth]{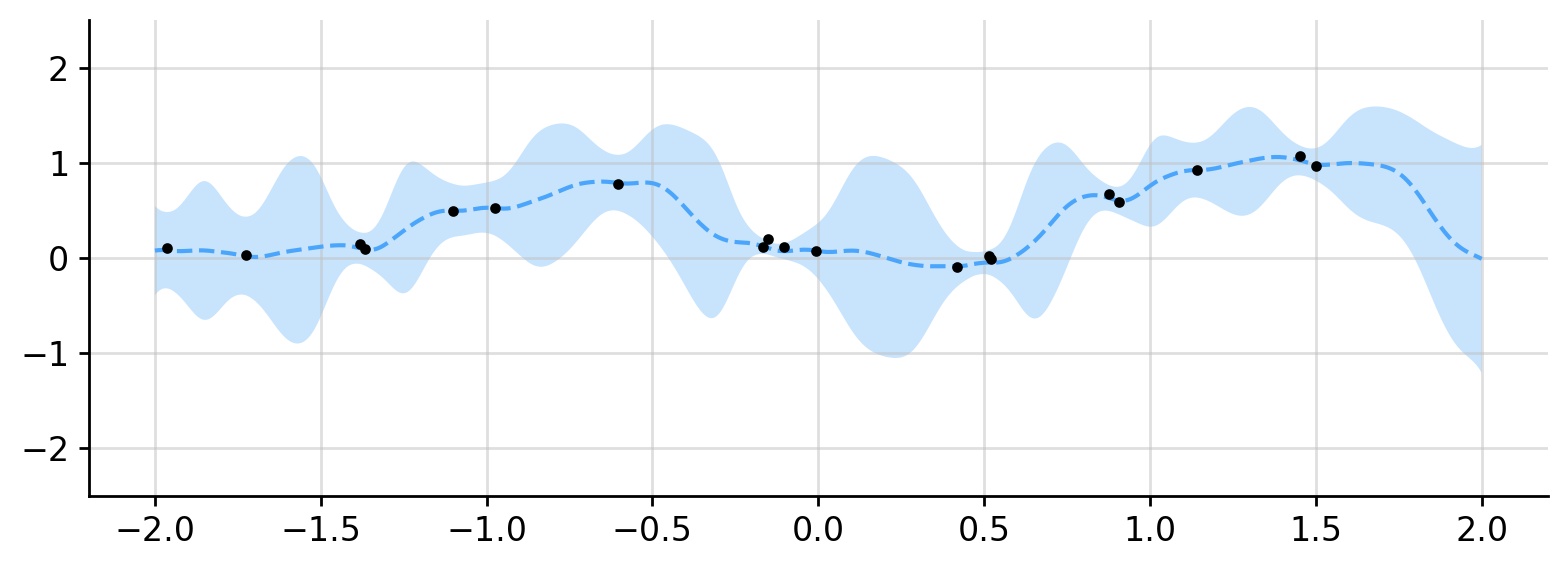}
    \caption{ConvCNP, $|\boldsymbol{\Xi}|=100$}
    \label{fig:subfig6}
  \end{subfigure}
  
  \caption{Meta-model predictions for SE covariance GP prior generated test dataset. Left column corresponds to the limited data regime, right column corresponds to the abundant data regime. Dashed lines represent predictive means, shaded regions represent 95\% confidence zones, black dots represent context datapoints.}
  \label{fig:se_toy_reg}
\end{figure}

For the first scenario, we train all meta-models on a meta-dataset consisting of SE covariance GP prior sample generated datasets, and visualise their predictions on a similarly generated but unseen dataset. This procedure is carried out for meta-dataset sizes $|\boldsymbol{\Xi}| \in \{1, 100\}$ to compare data-efficiency between the meta-models. The predictions are shown in Figure \ref{fig:se_toy_reg}.

\begin{figure}[h]
  \centering
  \begin{subfigure}{0.49\textwidth}
  \centering
    \includegraphics[width=\linewidth]{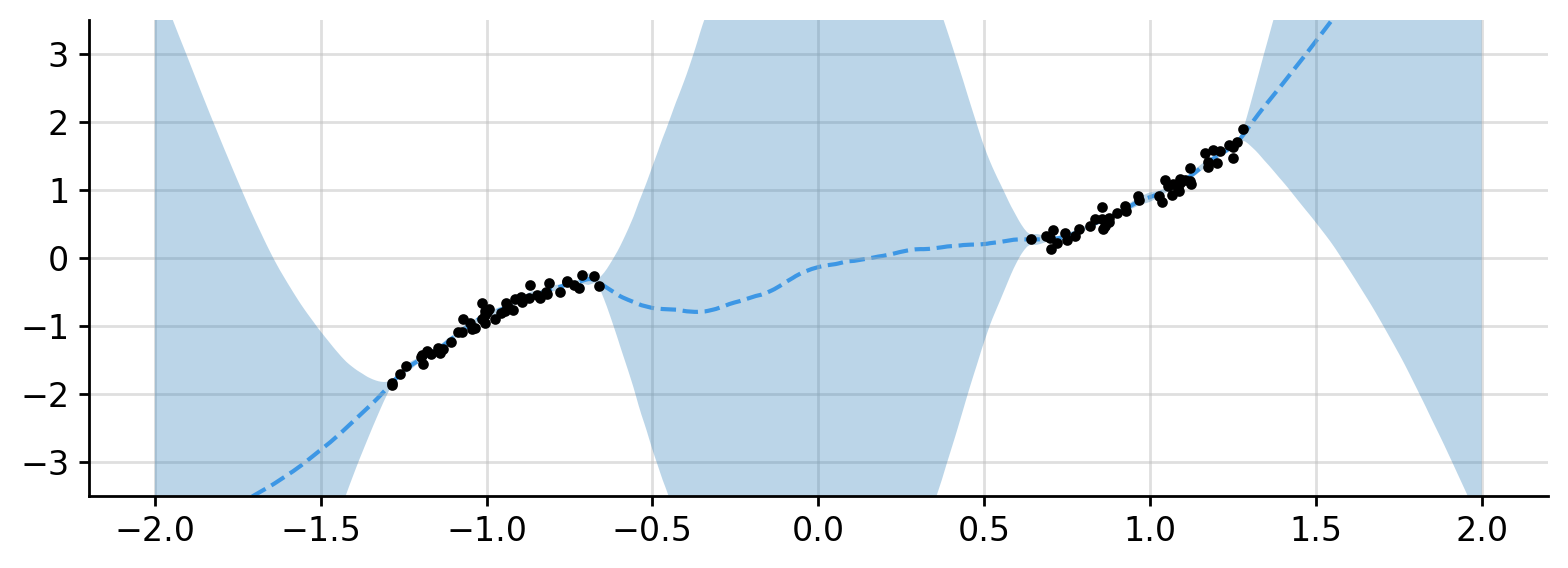}
    \caption{APOVI-BNN, $|\boldsymbol{\Xi}|=1$}
    \label{fig:subfig7}
  \end{subfigure}
  \begin{subfigure}{0.49\textwidth}
  \centering
    \includegraphics[width=\linewidth]{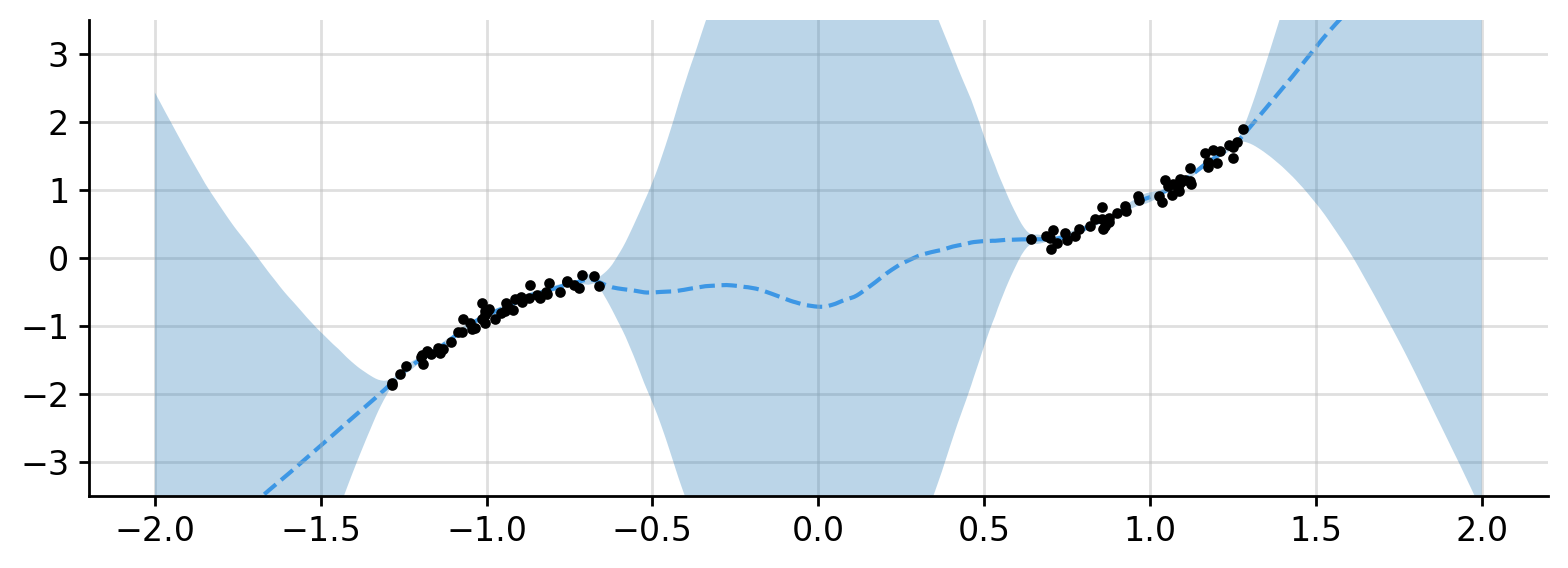}
    \caption{APOVI-BNN, $|\boldsymbol{\Xi}|=100$}
    \label{fig:subfig8}
  \end{subfigure}
  \vspace{0.5cm}
  
  \begin{subfigure}{0.49\textwidth}
  \centering
    \includegraphics[width=\linewidth]{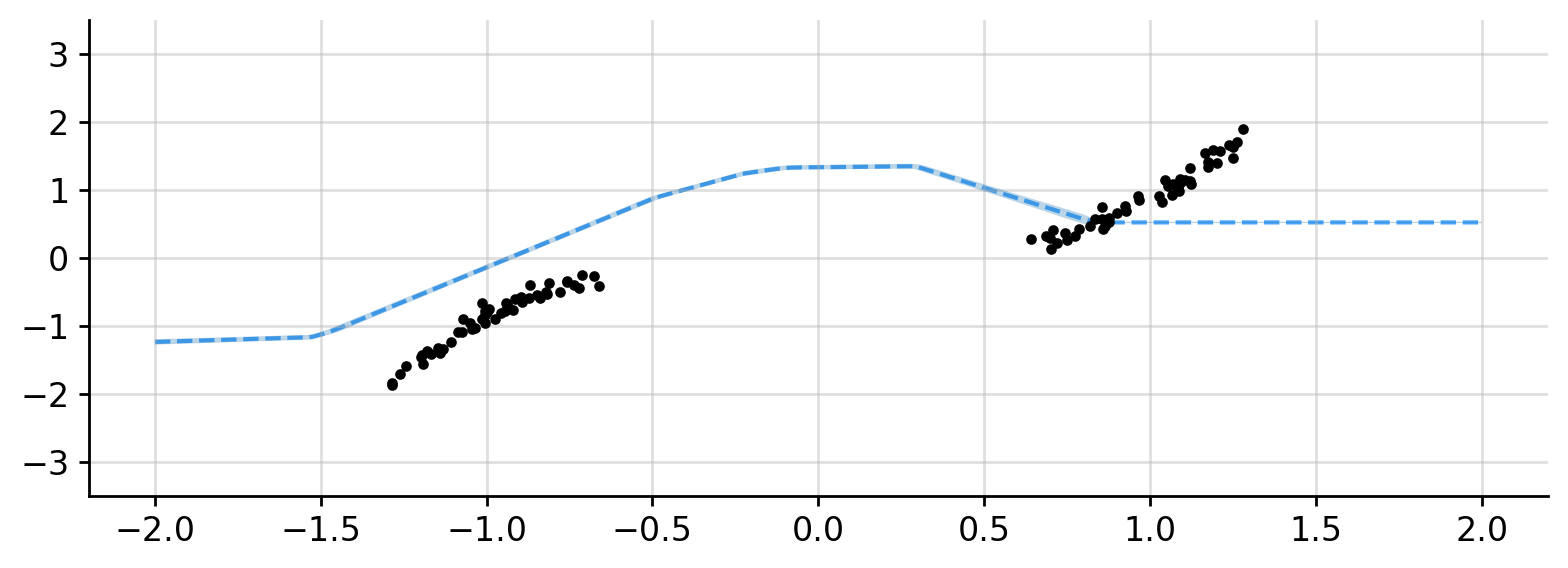}
    \caption{AMFVI-BNN, $|\boldsymbol{\Xi}|=1$}
    \label{fig:subfig9}
  \end{subfigure}
  \begin{subfigure}{0.49\textwidth}
  \centering
    \includegraphics[width=\linewidth]{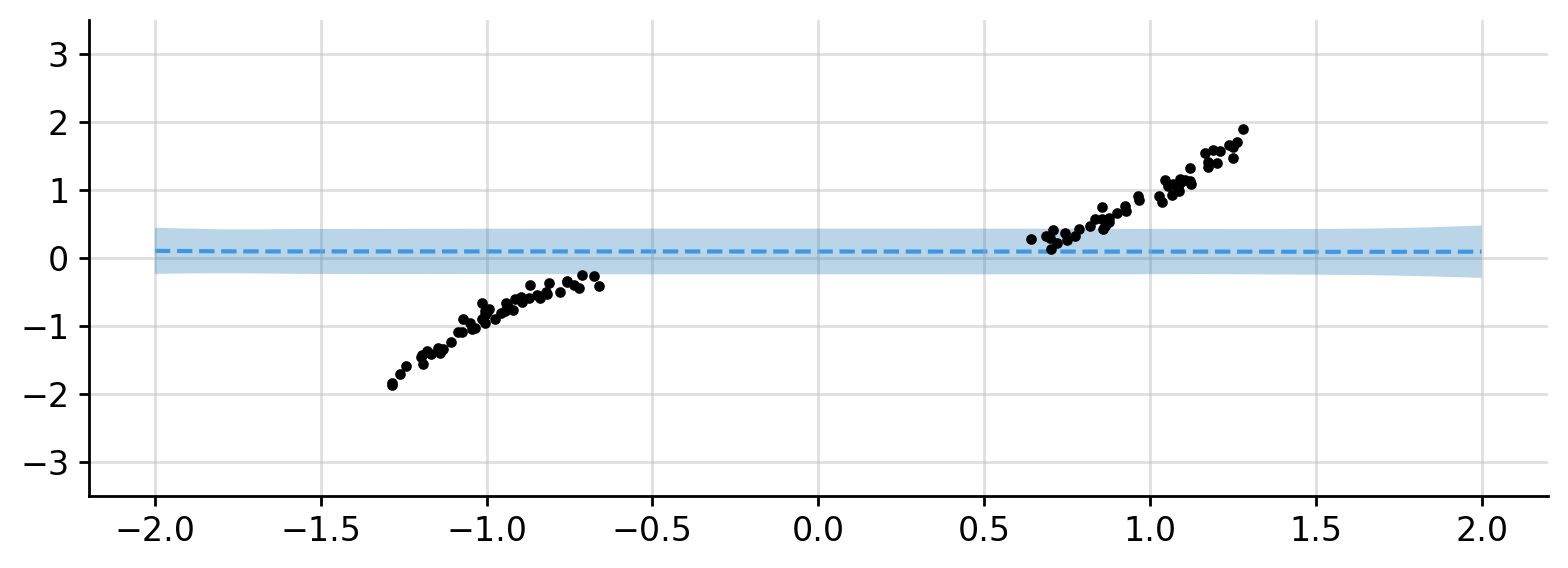}
    \caption{AMFVI-BNN, $|\boldsymbol{\Xi}|=100$}
    \label{fig:subfig10}
  \end{subfigure}
  \vspace{0.5cm}
  
  \begin{subfigure}{0.49\textwidth}
  \centering
    \includegraphics[width=\linewidth]{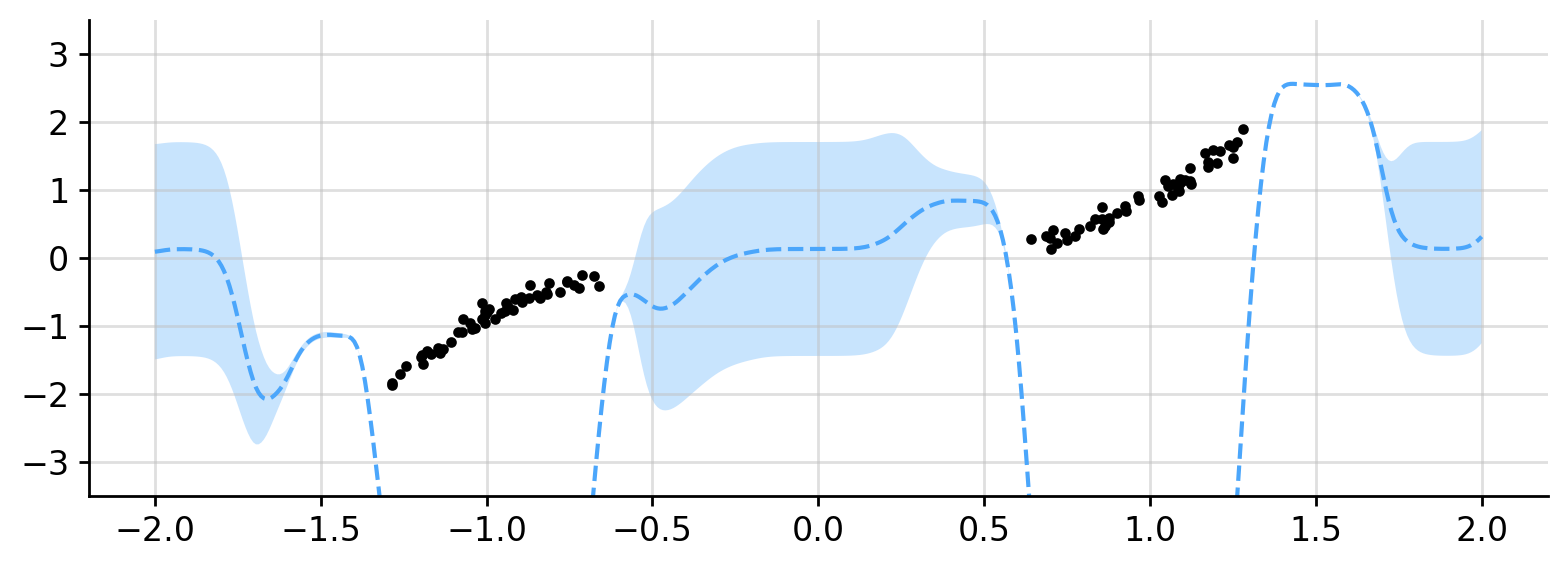}
    \caption{ConvCNP-BNN, $|\boldsymbol{\Xi}|=1$}
    \label{fig:subfig11}
  \end{subfigure}
  \begin{subfigure}{0.49\textwidth}
  \centering
    \includegraphics[width=\linewidth]{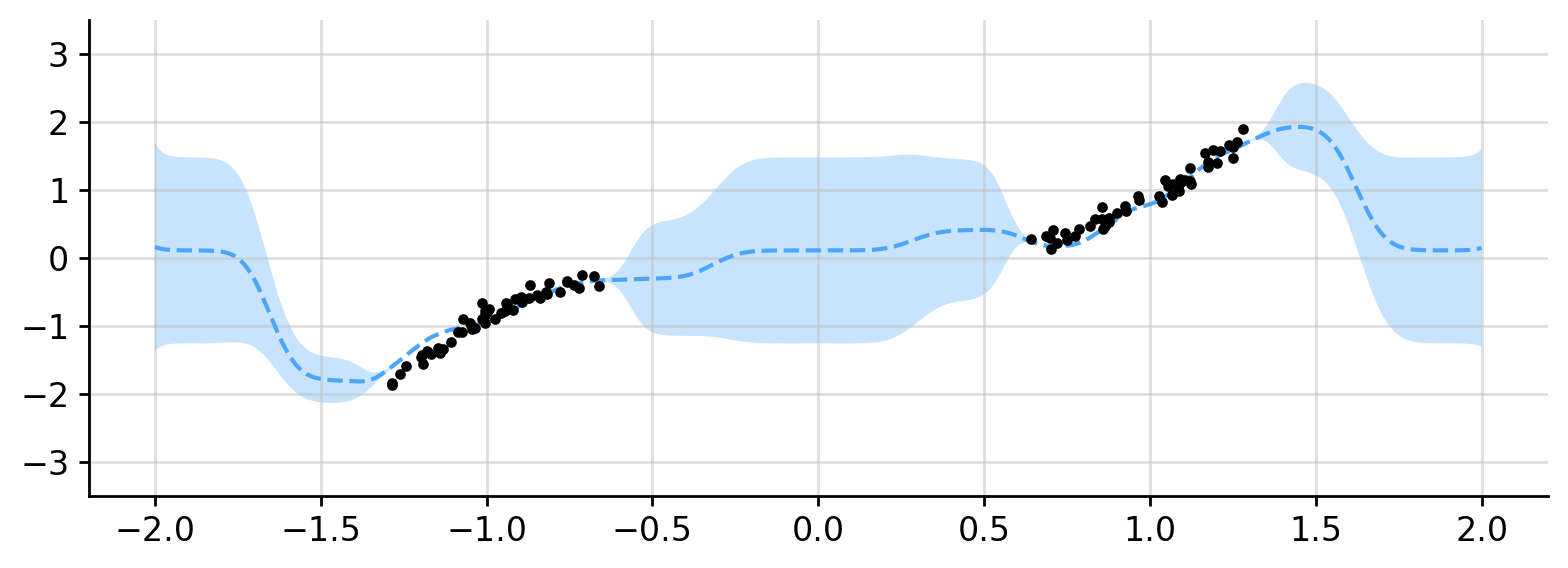}
    \caption{ConvCNP, $|\boldsymbol{\Xi}|=100$}
    \label{fig:subfig12}
  \end{subfigure}
  
  \caption{Meta-model predictions for cubic dataset. Left column corresponds to the limited data regime, right column corresponds to the abundant data regime. Dashed lines represent predictive means, shaded regions represent 95\% confidence zones, black dots represent context datapoints.}
  \label{fig:cubic_toy_reg}
\end{figure}

In the second scenario, we train the meta-models on similarly generated meta-datasets, but the meta-models are then evaluated on the cubic dataset from \cite{ober2021global}, which is quite different from an SE covariance GP function sample. This experiment is also repeated for meta-dataset sizes of $|\boldsymbol{\Xi}| \in \{1, 100\}$, and the obtained predictions are shown in Figure \ref{fig:cubic_toy_reg}

In both scenarios, the APOVI-BNN is the only model capable of producing sensible predictive distributions in the limited data setting. By contrast, the ConvCNP performs poorly in the limited data cases, exhibiting a pathology in which its predictive mean appears to \say{miss} the data despite reasonable variance behaviour. This affirms the belief that the ConvCNP relies on the many shared parameters overfitting to a large meta-dataset. In all cases, the AMFVI-BNN is the worst of the trio, and we see that when $|\boldsymbol{\Xi}| = 1$ its predictions are particularly confident, despite appearing to ignore the test datapoints entirely. What is particularly notable is the fact that the APOVI-BNN is the only model capable of producing rational predictions in the out-of-meta-dataset scenario, especially in the limited data regime. 

\begin{figure}[h]
    \centering
    \includegraphics[width=0.6\linewidth]{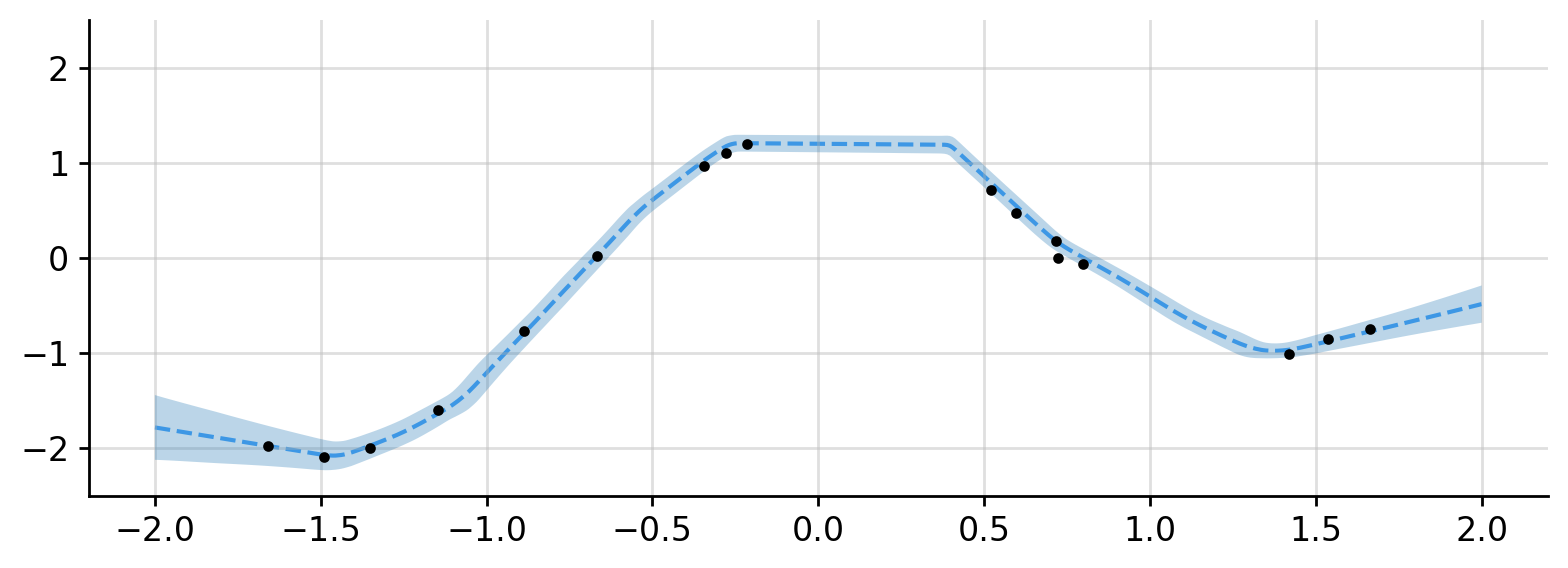}
    \caption{AMFVI-BNN predictions on the single SE covariance GP generated dataset on which it was trained}
    \label{fig:amfvi_train}
\end{figure}

To further explore the behaviour of the AMFVI-BNN, we visualise its predictions on the single dataset that was in the meta-dataset for the within-meta-dataset case in Figure \ref{fig:amfvi_train}, and note that the predictive distribution is near identical to that of Figure \ref{fig:subfig3}

\subsection{Image Completion}
\label{sec:img_comp}

The final way in which we evaluate the APOVI-BNN is in a two-dimensional regression setting in which datapoints correspond to pixels in an image. In this problem, pixels are randomly masked and the goal of the model is to predict the complete image. For a model such as an on-the-grid ConvCNP it is fairly straightforward to apply the model to this task, however for the APOVI-BNN we map pixel locations to a point $\mathbf{x} \in \mathbb{R}^2$ by setting the boundary pixels to be at locations $\pm1$ in the corresponding input dimension, and set the output $\mathbf{y}$ to be the pixel intensity. For greyscale images the dimensionality of $\mathbf{y}$ is $1$. This experiment was included in the original NP paper \citep{garnelo2018neural} to demonstrate the ability of NPs to \textit{learn} nontrivial \say{kernels}---a task that is an obvious limitation of GPs. We include it here to see if the APOVI-BNN can improve upon the ability of a ConvCNP to do this when the amount of data is limited.

We use the MNIST dataset \citep{726791} along with a mask that is generated randomly for each image to construct a meta-dataset of incomplete greyscale images, in which the pixel values are rounded to either 0 or 1 such that the problem can optionally be viewed as a pixel-wise binary classification problem. Each mask in the meta-dataset is produced by randomly selecting some probability $p\in[0.05, 0.95]$, and then assigning each pixel as unmasked with probability $p$ or masked with probability $1-p$ such that if $p=0.8$ then on average $80\%$ of the image pixels are left intact. For the ConvCNP, we use a heteroscedastic Gaussian likelihood in which the model learns both the mean and variance at every pixel location, while for the APOVI-BNN we employ a Bernoulli likelihood in which the model predicts the probability that a particular pixel takes the value 1. For full clarity, this means that the ConvCNP is performing regression while the APOVI-BNN is performing pixel-wise classification.

\begin{figure}[h]
    \centering
    
    \begin{subfigure}[b]{0.134\textwidth}
        \includegraphics[width=\textwidth]{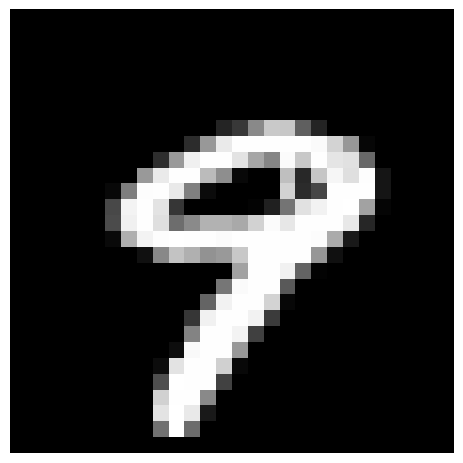}
        \caption{}
        \label{fig:image13}
    \end{subfigure}
    \begin{subfigure}[b]{0.134\textwidth}
        \includegraphics[width=\textwidth]{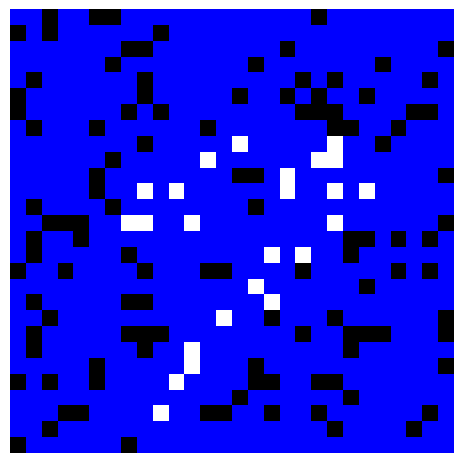}
        \caption{}
        \label{fig:image14}
    \end{subfigure}
    \begin{subfigure}[b]{0.134\textwidth}
        \includegraphics[width=\textwidth]{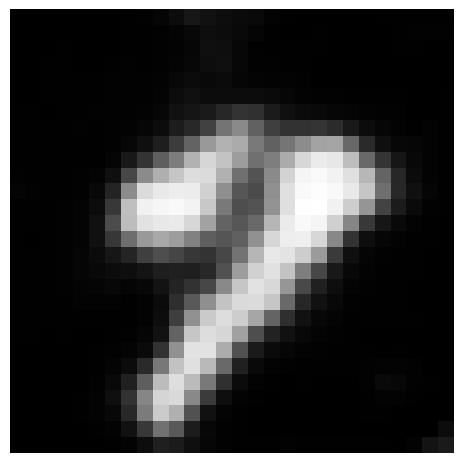}
        \caption{}
        \label{fig:image15}
    \end{subfigure}
    \begin{subfigure}[b]{0.134\textwidth}
        \includegraphics[width=\textwidth]{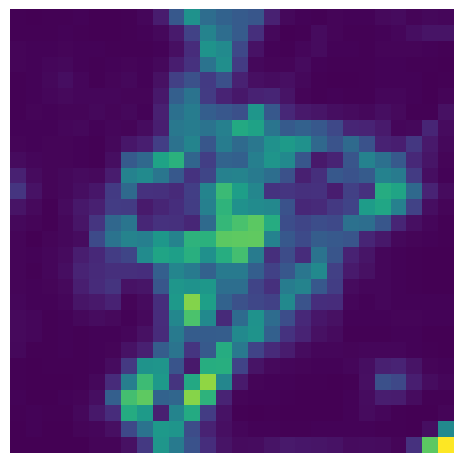}
        \caption{}
        \label{fig:image16}
    \end{subfigure}
    \begin{subfigure}[b]{0.134\textwidth}
        \includegraphics[width=\textwidth]{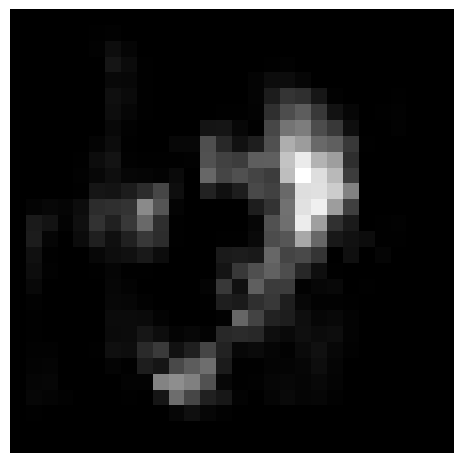}
        \caption{}
        \label{fig:image17}
    \end{subfigure}
    \begin{subfigure}[b]{0.134\textwidth}
        \includegraphics[width=\textwidth]{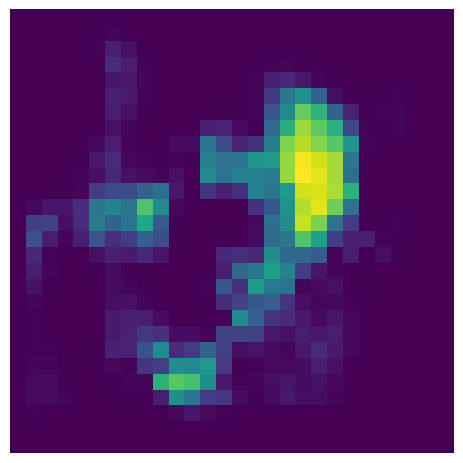}
        \caption{}
        \label{fig:image18}
    \end{subfigure}
    \begin{subfigure}[b]{0.134\textwidth}
        \includegraphics[width=\textwidth]{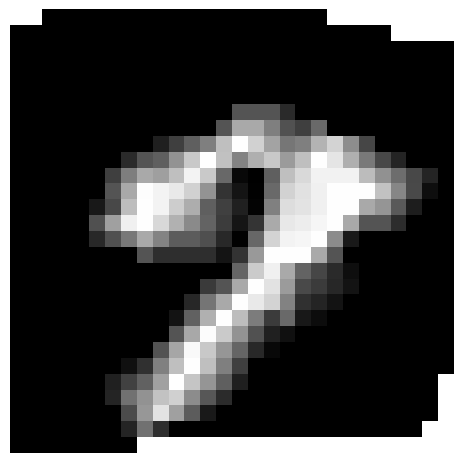}
        \caption{}
        \label{fig:image19}
    \end{subfigure}
    
    \caption{APOVI-BNN, ConvCNP, and linearly interpolated image completion predictions for an example masked MNIST image. From left to right the images depict \ref{fig:image13}: the original image, \ref{fig:image14}: the masked image, \ref{fig:image15}: the predictive mean of the APOVI-BNN, \ref{fig:image16}: the predictive standard deviation of the APOVI-BNN, \ref{fig:image17}: the predictive mean of the ConvCNP, \ref{fig:image18}: the predictive standard deviation of the ConvCNP, \ref{fig:image19}: a linear interpolation of the unmasked pixels. For these predictions each model was trained on a metadataset consisting of 10 datasets.}
    \label{fig:img_completion}
\end{figure}

We compute the squared error for the predictive mean of each model over 10 test images. We repeat the test in a scenario in which training data is abundant, where $|\boldsymbol{\Xi}|=60000$, as well as one in which data is limited, where $|\boldsymbol{\Xi}|=10$. The error is compared with that of a linear interpolator which is used as a baseline. The test images themselves, before and after masking, are the same across all tests. The masks of the test images are chosen with a probability $p$ randomly chosen from the [0.1, 0.3] interval. Note that 60,000 is chosen simply because it is the size of the MNIST training set. An example of the various types of prediction are shown in Figure \ref{fig:img_completion} for the limited data regime, and the squared error results\footnote{Note that the ConvCNP results corresponding to the abundant data setting are omitted. This is because convergence of the ConvCNP during training in this setting was highly troublesome. In the limited data setting, no such convergence issues were encountered.} are shown in Table \ref{tab:img_completion}.

\begin{table}[h]
    \centering
    \begin{tabular}{|l||r|r|}
        \cline{2-3}
        \multicolumn{1}{c||}{} & large $|\boldsymbol{\Xi}|$ & small $|\boldsymbol{\Xi}|$  \\
        \hline \hline
        APOVI-BNN & $20.6\pm1.3$ & $23.5\pm4.0$ \\
        \hline
        ConvCNP & - & $67.2\pm8.8$ \\
        \hline
        Lin. Interp. & $51.5\pm0.0$ & $51.5\pm0.0$ \\
        \hline
    \end{tabular}
    \caption{Sum of per-pixel squared error between original MNIST image and predicted completion of the masked image, averaged over 10 test images and 3 training repetitions.}
    \label{tab:img_completion}
\end{table}

The results in Table \ref{tab:img_completion} demonstrate that when the amount of data is limited, the ConvCNP is slightly outperformed by a linear interpolator, and significantly outperformed by the APOVI-BNN. The performance of the APOVI-BNN when there is little data is only slightly worse than when there is lots of data. Looking at a limited-data example in Figure \ref{fig:img_completion}, we see that the APOVI-BNN is capable of producing highly reasonable uncertainty estimates, while the ConvCNP is unable to do so with such little data.

\section{Discussion}

\textbf{Amortisation of the POVI-BNN.} Let us consider the implications of amortising the POVI-BNN in the way that we propose. One key assumption that is made is that the pseudo-likelihoods factorise across datapoints. Intuitively, this is saying that we expect the pseudo-datapoints in the pseudo-dataset for a given layer of the BNN to be independently distributed. Since this is the case for the actual dataset, which the pseudo-datasets are supposed to mimic, this assumption seems reasonable and so this is unlikely to significantly degrade performance compared to the POVI-BNN. 

Another major change is the exchanging of the set of learnable inducing locations for the actual data. This setup is in fact preferable to using inducing locations since there is more information about the dataset stored in the dataset itself than in a set of related inducing points, and so we expect the quality of the approximate posterior to be improved. The reason \citeauthor{ober2021global} use inducing points is to ensure that the POVI-BNN is scalable to large datasets---an application that the APOVI-BNN is not intended to be used for anyways. 

When the APOVI-BNN is used in a meta-learning setting in which the test dataset is only exposed to the model at test time, the model is only afforded a single forward pass in which to learn a good approximate posterior. It is in this setting that we expect the approximate posterior of the APOVI-BNN to be the most different to that of the POVI-BNN. 

The results of the ELBO experiment seem to be in agreement with the above points; when the APOVI-BNN is treated as a regular model, it performs better than the POVI-BNN, but when the test dataset is witheld until test time, the performance of the APOVI-BNN is degraded. The degree to which the APOVI-BNN surpasses the POVI-BNN in the $|\boldsymbol{\Xi}|=0$ and $|\boldsymbol{\Xi}|>2$ cases, however, is perhaps a little surprising. Seperately, when this experiment was performed it was noted that the APOVI-BNN took significantly less time to train than the POVI-BNN. A possible explanation for these two observations could be that the optimisation landscape for the amortised variant is more suited to gradient descent. Why exactly this is the case would need to be further investigated, but it is an explanation that seems to fit very well.

\textbf{Architecture Limitations.} In both the toy regression and image completion settings, the APOVI-BNN performs better than its competitors when the data is limited. However, its performance increases only marginally when the number of datasets is increased. This is likely because the APOVI-BNN is ultimately a simpler model than, say, a ConvCNP. The APOVI-BNN is limited by the fact that its predictive distributions are at best those of a Bayesian MLP; an MLP being a model that is not translationally invariant, not particularly scalable to large architectures when compared with architectures like CNNs, and not particularly suited to learning hierarchical representations of data. If we want the APOVI-BNN to perform as well as members of the NP family on complex tasks such as image completion, then the architecture needs to be \textit{very} large such that it stands a chance at learning complex hierarchies of representations, but then it becomes highly intensive to train since the complexity scales with network width cubed \citep{ober2021global}. There is the option of extending \citeauthor{ober2021global}'s convolutional POVI-BNN to the amortised setting as discussed, which would mean that, as long as the secondary inference networks are also convolutional, translation equivariance is baked in, but that model would also still only be computationally feasible for small datasets or images. 

With this limitation in mind, we can reason about the performance of the different models in the image completion experiment and why the APOVI-BNN performs so well in the limited data case despite using the NPML objective. When data is abundant, ConvCNPs learn to make predictions that are very similar to the training images. In the paper in which ConvCNPs are introduced \citep{gordon2020convolutional}, we see that if the ConvCNP is trained on MNIST, its predictions attempt to make a digit even if every pixel in the test image is masked; if the ConvCNP is trained on the CelebA facial dataset \citep{liu2015faceattributes}, then the model tries to construct a face even if every pixel in the test image is masked. For a well-trained ConvCNP, complex image structures that are shared by images in the training dataset are then ingrained into the predictions, and at test time the model only needs to use the available pixels to fine-tune the predictions. This is a sure sign of of overfitting to the meta-dataset; if we train the ConvCNP on images of faces and then test it on a masked image of a car, it will attempt to make a face and not a car since it is unable to generalise beyond the training meta-dataset. In the case of the APOVI-BNN in our image completion experiment, the BNN architecture consists of three hidden layers of 150 neurons---an architecture that is likely not large enough to learn such complex features. Instead, it seems that the APOVI-BNN only learns to perform a probabilistic interpolation at test time. This is a considerably more simple task, but one that requires significantly less overfitting and is much more generalisable as a result. We would argue that it is for this reason that, despite no longer being a Bayesian model under the NPML objective, the APOVI-BNN still exhibits high data efficiency in the image completion experiment.

\textbf{Choice of Objective.} The choice of training the APOVI-BNN with the ELBO, NPVI, or NPML training objectives was not a subject of particular focus in this project, but it is worth some consideration. Although use of the NPML objective surrenders the APOVI-BNN's status of being Bayesian, we still observe superior data-efficiency. As well as the possible architectural reason for this discussed above, this is likely helped by the fact that the mapping from data to the per-datapoint parameters of the distribution over weights is fairly simple and so not that much data is needed to learn it. We found that use of the NPVI objective in the image completion experiment left the APOVI-BNN unable to produce meaningful predictions. Perhaps one way to remain Bayesian and still produce good predictions would be to use a prior distribution that is less restrictive, but we leave an investigation of this to future work.

\textbf{AMFVI-BNN.} The performance of the AMFVI-BNN is poor in all cases. It was omitted from the image completion experiment since it was unable to produce meaningful predictions under any training objective. MFVI-BNNs are known to underfit the data \citep{dusenberry2020efficient}, and use of the amortised variant as a meta-model in the way we propose seems to worsen this pathology. Looking at Figures \ref{fig:subfig3} and \ref{fig:amfvi_train}, we see that the model overfits to the single training dataset and produces almost identical predictions on an unseen dataset; in this case it ignores the test data entirely. This is in stark contrast to the APOVI-BNN, which produces very sensible predictions on unseen datasets after exposure to a single dataset in training. The reason for the poor perfomance of the AMFVI-BNN is twofold. Firstly, the MFVI posterior approximation does not allow for dependencies between model weights to be modelled either within or across network layers. This is the reason for the underfitting pathology of MFVI-BNNs. Secondly, when used as a meta-model, the AMFVI-BNN is not flexible enough to learn to perform approximate Bayesian inference within a single forward pass. When there are many datasets on which to train, such as in Figure \ref{fig:subfig4}, we see the AMFVI-BNN learns to make predictions that ignore the data but that have wide enough uncertainty bounds that any dataset encountered can be explained by noise.

\section{Conclusion}

In this dissertation we presented a new probabilistic meta-model, the APOVI-BNN, that addresses the lack of data-efficiency shared by existing approaches. We derived the model by building upon \citeauthor{ober2021global}'s approximate posterior for BNNs, the POVI-BNN. We showed that, when used as a regular model, the APOVI-BNN is capable of producing superior approximate posterior distributions to \citeauthor{ober2021global}'s POVI-BNN. We discussed an alternative interpretation of the model; that it can be viewed as a neural process and that it is reasonable to train it with neural process training objectives in meta-learning settings as a result. Finally, we assessed our model as a probabilistic meta-model in both data-rich and data-scarce scenarios across one-dimensional artificial regression and image completion problems, and found that our method produces superior predictions to those of a ConvCNP when the data is limited. Along the way, we presented a second probabilistic meta-model that also builds upon a common variational approximate posterior for BNNs, the AMFVI-BNN, whose performance was significantly worse than that of the APOVI-BNN. This affirmed that the APOVI-BNN offers a particularly unique and effective solution to probabilistic meta-learning applications with limited data. 

\textbf{Future Work.} There are two main avenues which could be pursued for further research in the APOVI-BNN. The first is to conduct an empirical analysis of the behaviour of the APOVI-BNN under each of the three available training objectives; the ELBO, the NPML objective, and the NPVI objective. Although the ELBO and NPVI objectives are very similar, the NPVI incorporates a notion of rewarding good target datapoint predictions, and as a result it could result in better predictions while maintaining the model's Bayesian status. The second is to explore the performance of the APOVI-BNN on a real-world meta-dataset of practical consequence. Although much insight can be gained from performance evaluations on artificial regressions and small-scale image completions, the real-world applications of the model remain somewhat unexplored.

\section*{Acknowledgements}

I would like to acknowledge my supervisor Dr. Adrian Weller for his highly valuable comments and feedback throughout the year. I would also like to thank Matt Ashman for inviting me to pursue an idea of his as a project and for giving me so much of his time over the past year. He has provided me with invaluable advice both as a project co-supervisor and as a role-model in my academic career, and for everything I am immensely grateful.

I would also like to acknowledge Dr. Richard Turner, whose stunning lecture course \textit{3F8: Inference} introduced me to the beautiful world of probabilistic machine learning and some of the many jewels within. It is thanks to his excellent teaching that I developed such an interest in the field.

Finally, I would like to thank my physicist friends over at Portugal Place for countless stimulating discussions on and off the topic of this project, as well as my partner, Inca, for her loving support throughout the year.

\bibliography{bibliog}
\addcontentsline{toc}{section}{Bibliography}

\appendix
\section*{Appendix}

\section{Experimental details}
\label{app:exp}

\subsection*{ELBO Experiment}
Training and test datasets consisted of between 10 and 50 randomly selected inputs in the [-2.0, 2.0] interval. These locations were evaluated at functions sampled from SE covariance GP priors, with additive Gaussian noise of standard deviation $\sigma=0.05$ sprinkled on top to generate the corresponding outputs. 

Both the APOVI-BNN and POVI-BNN had two hidden layers of 32 neurons, and the secondary inference networks of the APOVI-BNN had two hidden layers of 50 neurons. For both models the observation noise at the output was fixed to 0.05. 5 test datasets that were generated with a constant seed were used to find an average ELBO value for each model. Training was initialised with a different seed over 5 repetitions for each model and training data combination; so for the APOVI-BNN each meta-dataset size required five repetitions, while for the POVI-BNN five repetitions were completed for each test dataset. The learning rate for the APOVI-BNN was $10^{-3}$ while for the POVI-BNN it was $5\times10^{-3}$. The maximum number of epochs for both models was 15,000 but early stopping was enabled after 1000 epochs with 500 epochs patience, with smoothing over 500 epochs. A batch size of 5 was used. ReLU activation functions were used throughout. The $\pm$ bound shown in the table is standard deviation.

\subsection*{1D Regressions}
Training datasets and the SE covariance GP test dataset consisted of between 10 and 20 points randomly selected from the [-2.0, 2.0] interval. These locations were evaluated at functions sampled from SE covariance GP priors, with additive Gaussian noise of standard deviation $\sigma=0.05$ sprinkled on top to generate the corresponding outputs. The cubic dataset was generated identically to in \cite{ober2021global}.

Both the APOVI-BNN and AMFVI-BNN had two hidden layers of 50 neurons and secondary inference networks also with two hidden layers of 50 neurons. For both models the observation noise was initialised at 0.01 but this parameter was trained. The ConvCNP used a discretisation of 125 points per unit, a functional embedding of dimension 32, a CNN with 16 channels in the first layer, 32 in the second, 16 in the final layer, a CNN kernel size of 11, and an initial lengthscale of 0.096, but this parameter was trained.

The A-BNNs were trained using the ELBO and a learning rate of $10^{-3}$. The maximum number of epochs was 15,000, but early stopping was enabled from 2000 epochs with 300 epochs patience and 500 epochs of smoothing. A batch size of 5 was used. The ConvCNP was trained with a learning rate of $3\times10^{-4}$. The maximum number of epochs for the ConvCNP was 10,000, but early stopping was enabled from 1000 epochs with 200 epochs patience and 200 epochs smoothing. A batch size of 5 was also used in the ConvCNP. Predictive distributions of A-BNNs were estimated by taking 100 samples from the predictive distribution and fitting a Gaussian with the same mean and variance to these samples.

\subsection*{Image Completion}
The data is prepared as detailed in Section \ref{sec:img_comp}. The APOVI-BNN used three hidden layers of 150 neurons and secondary inference networks with three hidden layers of 200 neurons. A Bernoulli likelihood was utilised as mentioned. The on-the-grid ConvCNP used a functional embedding of dimension 256, input and output convolutions with kernel size 5, and a CNN consisting of three convolutional layers of 256 channels with residual connections and a kernel size of 3. An MLP with a single hidden layer of width 128 was used to map from the final convolutions to the heteroscedastic likelihood parameters for each pixel.

The APOVI-BNN was trained using the NPML objective, a learning rate of $10^{-4}$, and a batch size of 4. The maximum number of epochs was 10,000, but early stopping was enabled from 500 epochs, with patience of 100 epochs and 100 epochs smoothing. The ConvCNP was trained with a learning rate of $3\times10^{-4}$ and a batch size of 10. The maximum number of epochs was 10,000, but early stopping was enabled from 600 epochs, with 100 epochs patience and 100 epochs smoothing.

\section{Implementation}

An implementation of the MFVI-BNN used for Figure \ref{fig:mfvi-bnn} can be found in \href{https://github.com/Sheev13/bnn-mean-field-vi}{this GitHub repository}. The codebase for the rest of the project can be found in \href{https://github.com/Sheev13/bnn_amort_inf}{the project's GitHub repository}. Note that regular git commits were submitted throughout the year, and so the repository's commit record serves as a project logbook.

The code used for this dissertation does not rely on any previously written software except for the standard python packages, notably including PyTorch and GPyTorch. Excluding GPs, which are mostly implemented already in GPyTorch, all models used were implemented from scratch. This includes implementations of:
\begin{enumerate}
    \item POVI-BNN
    \item MFVI-BNN
    \item APOVI-BNN
    \item AMFVI-BNN
    \item CNP
    \item on-the-grid ConvCNP
    \item off-the-grid ConvCNP
\end{enumerate}as well as secondary models that are used in one or more of the above:
\begin{enumerate}
    \item MLP
    \item CNN with or without residual connections
    \item Unet-style CNN
    \item SetConv
\end{enumerate}and finally partial implementations of GP's with the following kernels:
\begin{enumerate}
    \item SE
    \item periodic
    \item Laplacian.
\end{enumerate}Existing ConvCNP implementations were used to check the correctness of my ConvCNP implementations. These include an off-the-grid ConvCNP found in \href{https://github.com/cambridge-mlg/convcnp}{this} repository, and an on-the-grid ConvCNP found in \href{https://github.com/YannDubs/Neural-Process-Family}{this} one, but note that neither of these produced any plots used in this report.

Except for small changes such as bug fixes, refactors, or minor additions which were completed by Matthew Ashman, all code in the repository was written by myself. Full details can be found in the repository's commit history. Note that there are some files in the repository written entirely by Matt---these were used for paper submissions based on this project but not for this report.
\\ \\ \\ \\

\begin{center}
    \textbf{Wordcount: 9933}
\end{center}

\end{document}